\documentclass[lettersize,journal]{IEEEtran}
\usepackage{amsmath,amsfonts}
\usepackage{algorithmic}
\usepackage{algorithm}
\usepackage{array}
\usepackage[caption=false,font=normalsize,labelfont=sf,textfont=sf]{subfig}
\usepackage{textcomp}
\usepackage{stfloats}
\usepackage{url}
\usepackage{verbatim}
\usepackage{graphicx}
\usepackage{cite}
\usepackage{graphicx}
\usepackage{amsmath}
\usepackage{amssymb}
\usepackage{booktabs}
\usepackage{bm}
\usepackage{multirow} 
\usepackage{algorithm}
\usepackage{algorithmic}
\usepackage{enumitem}
\usepackage{siunitx}
\begin{document}
	
	\title{CIRKDV2: Cross-Image Relational Knowledge Distillation with Contextual Modeling for Efficient Semantic Segmentation}
	
	\author{Chuanguang Yang, Yu Wang, Chengqing Yu, Xinqiang Yu, Weilun Feng, Yuqi Li, Zhulin An, Libo Huang, Boyu Diao, Fei Wang, Fuzhen Zhuang,~\IEEEmembership{Senior Member,~IEEE}, Yongjun Xu, Yingli Tian,~\IEEEmembership{Fellow,~IEEE}, Tingwen Huang,~\IEEEmembership{Fellow,~IEEE}, and Yongduan Song,~\IEEEmembership{Fellow,~IEEE}
		\thanks{Chuanguang Yang, Yuqi Li, Zhulin An, Libo Huang, Boyu Diao, Fei Wang, and Yongjun Xu are with the Institute of Computing Technology, Chinese Academy of Sciences, Beijing 100190, China (e-mail: yangchuanguang@ict.ac.cn; yuqili010602@gmail.com; anzhulin@ict.ac.cn; www.huanglibo@gmail.com; diaoboyu2012@ict.ac.cn; wangfei@ict.ac.cn; xyj@ict.ac.cn).}
		\thanks{Yu Wang, Chengqing Yu, and Weilun Feng are with the Institute of Computing Technology, Chinese Academy of Sciences, Beijing 100190, China, and also with University of Chinese Academy of Sciences, Beijing 100049, China (e-mail: wangyu23s@ict.ac.cn; yuchengqing@ict.ac.cn; fengweilun24s@ict.ac.cn).}
		\thanks{Xinqiang Yu is with the Institute of Automation, Chinese Academy of Sciences, Beijing 100190, China, and also with University of Chinese Academy of Sciences, Beijing 100049, China (e-mail: xinqiang.yu@nlpr.ia.ac.cn).}
		\thanks{Fuzhen Zhuang is with Institute of Artificial Intelligence, Beihang University, Beijing, China (e-mail: zhuangfuzhen@buaa.edu.cn).}
		\thanks{Yingli Tian is with the Department of Electrical Engineering, The City College, and the Department of Computer Science, the Graduate Center, the City University of New York, New York, NY, 10031 (e-mail: ytian@ccny.cuny.edu).}
		\thanks{Tingwen Huang is with the School of Computer Science and Control Engineering, Shenzhen University of Advanced Technology, Shenzhen 518107, China (e-mail: huangtingwen@suat-sz.edu.cn).}
		\thanks{Yongduan Song is with the School of Automation, Chongqing University, Chongqing 400044, China (e-mail: ydsong@cqu.edu.cn).}
		\thanks{Corresponding author: Zhulin An.}}

	
	
	
	\maketitle
	
	\begin{abstract}
		Existing Knowledge Distillation (KD) methods for semantic segmentation often guide the student to mimic the teacher's structured information generated from an individual image. However, they ignore the global semantic relations across various images that are valuable for KD. This paper proposes Cross-Image Relational KD V2 (CIRKDV2), which focuses on transferring structured pixel-to-pixel, pixel-to-region, and channel-to-channel relations modeled from global images in the dataset. A large teacher network often constructs a well-structured feature space in terms of global contextual dependencies. CIRKDV2 makes the student mimic the better structured semantic relations of the teacher, thus improving the feature representational capability. Beyond relational distillation, we further propose contextual feature distillation by using context aggregation to augment student pixel embeddings to match the semantic level with the teacher. This avoids the semantic gap problem between the teacher and student occurred in traditional feature distillation studies. Experimental results on Cityscapes, ADE20K, Pascal VOC, and COCO-Stuff-164K demonstrate that CIRKDV2 outperforms the state-of-the-art semantic segmentation distillation methods by large margins across various network architectures. Our code and pre-trained models are publicly available at https://github.com/winycg/CIRKD.
	\end{abstract}
	
	
	\section{Introduction}
	Semantic segmentation is a crucial and challenging task in computer vision. It aims to classify each pixel in the input image with an individual category label. The applications of segmentation often focus on autonomous driving, virtual reality, and robot navigation. Although popular state-of-the-art segmentation networks, such as DeepLab~\cite{chen2017deeplab,chen2018encoder}, PSPNet~\cite{zhao2017pyramid}, and SAM~\cite{kirillov2023segment}, achieve remarkable performance, they are often equipped with large image encoders, leading to high computational costs. This weakness makes them difficult to deploy for real-world scenarios on resource-limited mobile devices. Model compression is an effective way to pursue compact networks, mainly divided into quantization~\cite{feng2025q,feng2026s,feng2026mpq}, pruning~\cite{yang2019multi,cai2022prior}, and knowledge distillation (KD)~\cite{hinton2015distilling,yang2021hierarchical,yang2023online}.

	\begin{figure}[!t]
		\centering
		\subfloat[Intra-image KD]{\includegraphics[width=0.22\textwidth]{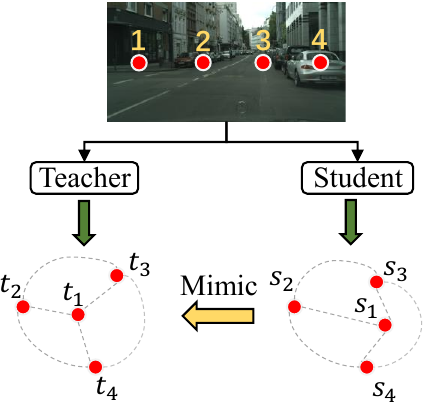}%
			\label{intra}}
		\hfil
		\subfloat[Cross-image KD]{\includegraphics[width=0.252\textwidth]{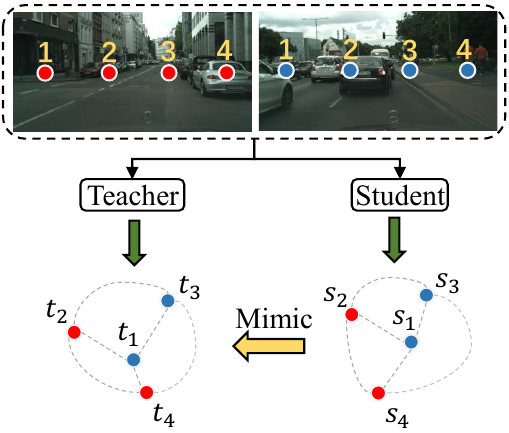}%
			\label{cross}}
		\caption{Overview of intra-image (\emph{left}) and our proposed cross-image relational distillation (\emph{right}). The circles with the same color denote pixel embeddings from the same image. $t_{i}$ and $s_{i}$ represent the pixel embeddings of the $i$-th pixel location tagged in an image from the teacher and student, respectively. The dotted line shows the similarity relationship between two pixels. The circles and lines construct a relational graph.} 
		\label{introduction}
	\end{figure}

	This paper investigates KD to improve the performance of a compact student network under the guidance of a high-capacity teacher network for semantic segmentation. A broad range of KD approaches~\cite{hinton2015distilling,yang2022mixskd,yang2025mtkd}  have been well studied but mostly for image classification tasks. Unlike image-level recognition, the segmentation task aims at dense pixel predictions, which is more challenging. Previous researches~\cite{liu2020structured} have found that directly utilizing classification-based KD methods to deal with dense prediction tasks may not achieve desirable performance. This is because strictly aligning the coarse feature maps between the teacher and student networks may lead to negative constraints and ignore the structured context among pixels.
	
	Recent studies attempt to propose specialized KD methods~\cite{xie2018improving,liu2019structured,he2019knowledge,wang2020intra,shu2021channel,tian2022adaptive,fan2023augmentation,liu2024rethinking} for semantic segmentation, mainly divided into relation- and feature-based distillation methods. Most relation-based distillation studies focus on mining correlations or dependencies among spatial pixel locations because segmentation needs a structured output. Typical knowledge can be local pixel affinity~\cite{xie2018improving,tian2022adaptive}, global pairwise relations~\cite{liu2019structured,he2019knowledge,fan2023augmentation} and intra-class pixel variation~\cite{wang2020intra}. Such methods often perform better than traditional point-wise alignment in capturing structured spatial knowledge. However, previous segmentation KD methods often guide a student to mimic the teacher's structured information generated from \emph{an individual image sample}. They ignore cross-image semantic relations among pixels for knowledge transfer, as shown in Figure~\ref{introduction}. Therefore, \textbf{they only construct narrow context but cannot capture wide contextual dependencies across global images.}

	To address this problem, we propose Cross-Image Relational Knowledge Distillation V2 (CIRKDV2) for semantic segmentation. To capture wide contextual dependencies, we aim to construct global relations across the whole training images from both the teacher and student. A good pre-trained teacher network with high performance could often generate a well-structured embedding space and capture better correlations than a student network. Based on this property, we distill pixel-to-pixel, pixel-to-region, and channel-to-channel relations across various images from the teacher to the student. This helps the student fully explore multi-dimensional structured relations to learn long-range contextual dependencies. The pixel-to-pixel and channel-to-channel distillation aims to transfer similarity distributions among pixel and channel embeddings, respectively. The channel embedding means a semantic map along the channel dimension of a feature map. The pixel-to-region distillation focuses on transferring pixel-to-region similarity distributions between pixel and region embeddings. The region embedding is generated by averagely pooling pixel embeddings from the same class and represents the feature center of that class. The pixel-to-region relations indicate the relative similarities between pixels and class-wise prototypes.
	
	A naive way for constructing cross-image relations is to derive embeddings from the current mini-batch. However, the batch size of the segmentation task is often small, limiting the network to capture broader pixel dependencies. Motivated by the memory mechanism in previous self-supervised learning~\cite{wu2018unsupervised}, we introduce a pixel queue, a region queue, and a channel queue in the memory bank to store abundant embeddings for modeling long-range relations. The embeddings in the queue are \textit{consistent in distribution} during the distillation process, since they are generated from the pre-trained and frozen teacher network. We regard the teacher and student pixel embeddings from the current mini-batch as anchors, and then randomly sample contrastive pixel and region embeddings from queues to model pixel-to-pixel as well as pixel-to-region similarity distributions. In a similar manner, we regard channel embeddings from the current mini-batch as anchors, and then randomly sample channel embeddings from the channel queue to construct channel-to-channel similarity distributions. Then we align such similarity distributions via KL-divergence from the student to the teacher. This helps the student learn comprehensive structured relation knowledge from the teacher.
	
	Beyond popular relation-based distillation, feature-based distillation is also demonstrated as an important technique. Shu \emph{et al.}~\cite{shu2021channel} revealed that each channel represents a category-specific mask and thus proposed Channel-Wise Distillation (CWD). CWD demonstrates the importance of channel-level information for dense prediction tasks. More recently, LAD~\cite{liu2024rethinking} transfers the classical FitNet idea to semantic segmentation for raw feature distillation. \textbf{However, previous methods often perform direct alignment without contextual modeling for student pixels, leading to possible mismatching semantics between the teacher and student.} This is because the teacher is often equipped with a larger kernel than the student,  which means that the teacher pixel extracts semantic information from a wider receptive field. To mitigate the semantic gap, we propose context aggregation over the spatial dimension of the student feature map to augment pixel embeddings with wider contextual information. And then we perform feature distillation of pixel embeddings between the student and teacher. This helps the student learn meaningful pixel semantics from the teacher without negative receptive field gaps.

	In summary, CIRKDV2 leverages pixel-to-pixel, pixel-to-region, and channel-to-channel dimensions for cross-image relational distillation, and contextual feature distillation to improve the student's segmentation performance under the guidance of a strong teacher. We evaluate our method over popular DeepLabV3~\cite{chen2018encoder}, PSPNet~\cite{zhao2017pyramid}, UperNet~\cite{xiao2018unified}, and SegFormer~\cite{xie2021segformer} architectures with various backbones on four challenging semantic segmentation benchmark datasets: Cityscapes~\cite{cordts2016cityscapes}, ADE20K~\cite{zhou2017scene}, Pascal VOC~\cite{everingham2010pascal}, and COCO-Stuff-164K~\cite{caesar2018coco}. Experimental results indicate that CIRKDV2 achieves the best performance among the state-of-the-art distillation approaches, demonstrating the effectiveness of cross-image relational distillation and contextual feature distillation in semantic segmentation.
	
	Parts of this paper, including pixel-to-pixel and pixel-to-region distillation, were published originally in our conference version~\cite{yang2022cross}. This paper extends our earlier work in several valuable aspects for relation and feature distillation. The main contributions are summarized as follows:
	\begin{itemize}[noitemsep,nolistsep,,topsep=0pt,parsep=0pt,partopsep=0pt]
		\item We propose cross-image relational KD by modeling pixel-to-pixel, pixel-to-region, and channel-to-channel relations assisted by a memory bank mechanism for structured knowledge transfer. This helps the student learn comprehensive long-range semantic dependencies of various object segments across the whole images.
		\item We propose contextual feature distillation by using context aggregation to augment student pixel embeddings with wider contextual information and then performing feature alignment between the teacher and student. This also avoids the semantic mismatching problem occurred in traditional feature distillation studies.
		\item CIRKDV2 achieves the best performance among the state-of-the-art distillation methods across various architectures on four challenging semantic segmentation datasets.
	\end{itemize}
	
	\section{Related Work}
	\subsection{Knowledge Distillation for Image Classification}
	The core idea of KD~\cite{hinton2015distilling} is to transfer meaningful knowledge from a cumbersome teacher into a smaller and faster student. Most current KD methods deal with image classification networks, mainly divided into probability-based, feature-based, and relation-based approaches. Probability-based KD~\cite{hinton2015distilling} transfers the class probabilities produced by the teacher as soft labels to supervise the student. Feature-based KD focuses on intermediate feature maps~\cite{romero2014fitnets} or their refined information~\cite{zagoruyko2016paying} as knowledge. Relation-based KD~\cite{fang2021seed,yang2022mutual,feng2024relational} aligns correlations or dependencies among multiple instances between the student and teacher networks. However, these image-level KD methods are often unsuitable for pixel-wise semantic segmentation~\cite{liu2020structured}, because the latter needs more fine-grained structured knowledge.
	\subsection{Relation-based Distillation for Semantic Segmentation}
	Existing KD methods for semantic segmentation can be divided into relation-based and feature-based approaches, where relation KD has become a mainstream paradigm. The core idea of relation-based KD is to encode contextual pixel or channel affinity as valuable knowledge for transfer. The main differences among existing work lie in how to construct relationships. Xie \emph{et al.}~\cite{xie2018improving} align local similarity maps constructed from 8 neighborhood pixels between the student and teacher networks. He \emph{et al.}~\cite{he2019knowledge} transfer non-local pairwise affinity maps using an autoencoder to minimize the discrepancy of features. SKD~\cite{liu2019structured,liu2020structured} performs pairwise similarity distillation among pixels and an adversarial distillation of score maps. IFVD~\cite{wang2020intra} distills intra-class feature variation to learn more robust relations with class-wise prototypes. DSD~\cite{feng2021double} constructs pixel-wise and category-wise similarity distillation, where the former uses residual attention maps to strengthen fine-grained spatial relations, and the latter aims to learn category correlation. APD~\cite{tian2022adaptive} mines sample-wise contextual information by an adaptive local perspective.
	Af-DCD~\cite{fan2023augmentation} distills dense structured relations from channel and spatial dimensions primarily within \textbf{the current image or mini-batch feature structure}. CIRKDV2 changes the knowledge being transferred: it models channel-to-channel relations across global images by using teacher channel embeddings stored in an online memory queue. Thus, each current student channel is not only compared with channels from the same image, but also with a broader set of teacher channel embeddings collected from previous images. This transfers dataset-level channel correlation patterns that are not available from an individual image alone.
	
	Although the above relation-based distillation methods achieve good performance, their common weakness is that the relational modeling is only limited in an individual image, which cannot capture long-range pixel dependencies. By contrast, CIRKDV2 proposes pixel-to-pixel and pixel-to-region distillation by sampling pixels from an online memory bank. Compared to the mini-batch, the memory bank significantly increases the number of available pixels for affinity modeling. Therefore, the student can learn longer-range pixel dependencies.
	
	\subsection{Feature-based Distillation for Semantic Segmentation}
	Beyond relation-based distillation, some studies examine feature-based distillation to improve the representational capability of the student network. Different from affinity modeling along the spatial dimension, CWD~\cite{shu2021channel} conducts channel-wise distillation to guide each student channel to mimic the corresponding teacher channel, where a channel often represents a semantic mask. SSTKD~\cite{ji2022structural,ji2025structural} distills structural and statistical texture features to let the student learn low-level detail information from the teacher. Attention distillation~\cite{zagoruyko2016paying} has been widely used for image classification. The attention map~\cite{woo2018cbam} often emphasizes meaningful semantic information while ignoring meaningless regions. The core idea of attention distillation is to make the student learn critical semantic information by mimicking the teacher's attention maps. This idea is also extended to semantic segmentation. CAT~\cite{cho2022class} distills class attention maps to guide the student to focus on class-discriminative regions of the teacher. AttnFD~\cite{mansourian2024attention} introduces CBAM~\cite{woo2018cbam} to extract channel and spatial attention maps and performs alignment between teacher and student. LAD~\cite{liu2024rethinking} revisits the seminal work FitNet~\cite{romero2014fitnets} and conducts feature distillation from $l_2$-normalized raw features. 
	
	Being orthogonal to relation-based distillation, feature-based distillation is also a complementary way to improve the student network. Previous feature-based methods often perform simple alignment between feature maps without high-level feature modeling. By contrast, we propose contextual modeling to augment feature representations, facilitating knowledge transfer of valuable high-level semantic information.
	
	\section{Methodology}
	\subsection{Preliminary}
	\textbf{Notations of Semantic Segmentation Framework.} Unlike traditional image classification, semantic segmentation is a pixel-wise dense classification task. A segmentation network needs to classify each pixel in the image to an individual category label from $C$ classes. The network can be decomposed of a feature extractor and a classifier. The former generates a dense feature map $\mathbf{F}\in \mathbb{R}^{H\times W\times D}$, where $H$, $W$, and $D$ denote the height, width, and number of channels, respectively. We can derive $H\times W$ pixel embeddings along the spatial dimension or $D$ channels along the channel dimension. The latter classifier further transforms the $\mathbf{F}$ into a categorical logit map $\mathbf{Z}\in \mathbb{R}^{H\times W\times C}$. The conventional segmentation task loss is to train each pixel with its ground-truth label using cross-entropy:
	\begin{equation}
		L_{task}=\frac{1}{H\times W}\sum_{h=1}^{H}\sum_{w=1}^{W} CE(\sigma(\mathbf{Z}_{h,w}),y_{h,w}).
		\label{task}
	\end{equation}
	Here, $CE$ denotes the cross-entropy loss, $\sigma$ denotes the softmax function and $y_{h,w}$ denotes the ground-truth label of the $(h,w)$-th pixel.
	
	\textbf{Pixel-wise Class Probability Distillation.} Motivated by Hinton's KD~\cite{hinton2015distilling}, a direct method is to align the class probability distribution of each pixel from the student to the teacher. The formulation is expressed as:
	\begin{equation}
		L_{kd}=\frac{1}{H\times W}\sum_{h=1}^{H}\sum_{w=1}^{W} KL(\sigma(\frac{\mathbf{Z}^{s}_{h,w}}{T})||\sigma(\frac{\mathbf{Z}^{t}_{h,w}}{T})).
		\label{kd}
	\end{equation}
	Here, $\sigma(\mathbf{Z}^{s}_{h,w}/T)$ and $\sigma(\mathbf{Z}^{t}_{h,w}/T)$ represent the soft class probabilities of the $(h,w)$-th pixel produced by the student and teacher, respectively.  $KL$ denotes the Kullback-Leibler divergence, and $T$ is a temperature. Following previous studies~\cite{liu2019structured,wang2020intra}, $T=1$ is good enough.

	\subsection{Overview of CIRKDV2}
\begin{figure*}[tbp]  
		\centering 
		\includegraphics[width=1\linewidth]{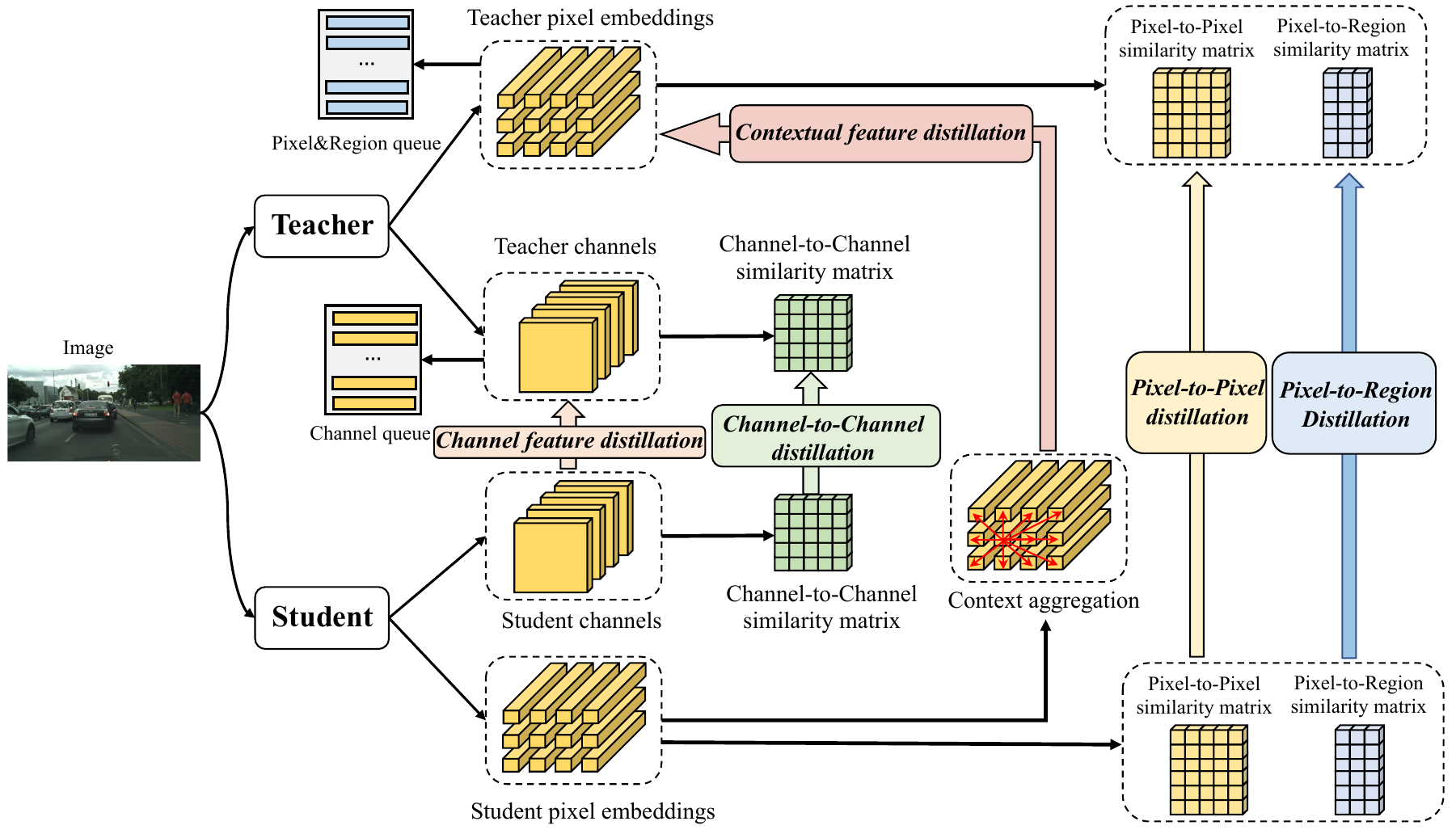}
		\caption{Overview of our proposed CIRKDV2.}
		\label{Overview_CIRKDV2}
	\end{figure*}

	A unified framework is shown in Figure \ref{Overview_CIRKDV2}, which illustrates the complete pipeline of CIRKDV2. Compared to the previous CIRKD, CIRKDV2 is an omni-dimensional distillation framework, including not only pixel-to-pixel distillation and pixel-to-region distillation, but also channel-to-channel distillation, channel feature distillation and contextual feature distillation.
	
	Semantic segmentation requires both spatially structured prediction and channel-wise semantic discrimination. Pixel-to-pixel distillation captures long-range spatial dependencies among pixels. Pixel-to-region distillation captures the relation between pixel embeddings and class-wise semantic prototypes. Channel-to-channel distillation captures the correlations among semantic response maps. These three dimensions are complementary: the first two focus on spatial/category-level structure, while the third focuses on semantic-channel interaction.
	
	Cross-image relational distillation transfers structural knowledge by aligning similarity distributions. However, relation losses mainly constrain relative similarities, not the absolute semantic content of each pixel embedding. Contextual feature distillation complements this by providing a direct teacher-guided semantic alignment after context aggregation. Therefore, it serves as a semantic anchor for the relational framework.

	\subsection{Pixel-wise Cross-Image Relational Knowledge Distillation}
	\textbf{Motivation.} Although the training objectives of $L_{task}$ and $L_{kd}$ are widely used in semantic segmentation, they only deal with pixel-wise predictions independently but neglect semantic relations between pixels. Some segmentation KD methods~\cite{liu2019structured,he2019knowledge,wang2020intra} attempt to capture spatial  relational knowledge by modeling pixel affinity. Nevertheless, these KD methods
	only construct the relationships among pixels \emph{within a single image}, regardless of the semantic dependencies among pixels across global images.  This paper demonstrates that cross-image relational knowledge is also valuable for conducting teacher-student-based KD.
	
	Our CIRKD makes use of pixel embeddings beyond a single image. Inspired by the recent memory-based contrastive learning~\cite{wu2018unsupervised,chen2020simple}, we may retrieve pixel embeddings of other images from \emph{the current mini-batch} or \emph{an online memory bank}. This paper considers both of two manners to model relationships among pixels, the details of which are shown as follows.

\subsubsection{Mini-batch-based pixel-to-pixel distillation}
	
	\label{minibatch}

\begin{figure*}[!t]
		\centering
		\subfloat[Mini-batch pixel-to-pixel distillation]{\includegraphics[width=0.36\textwidth]{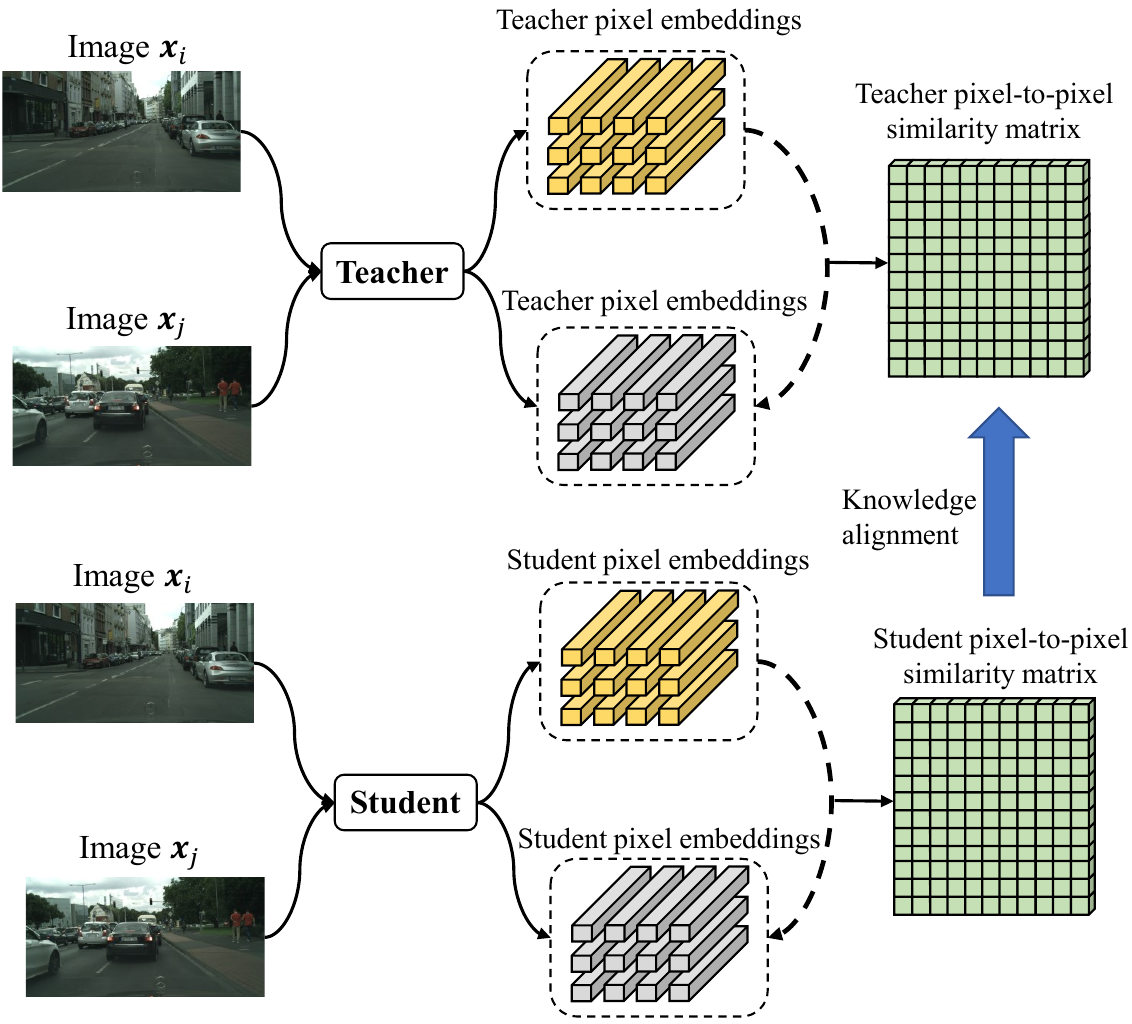}%
			\label{mini_pixel_kd}}
		\subfloat[Memory-based pixel-to-pixel and pixel-to-region distillation]{\includegraphics[width=0.64\textwidth]{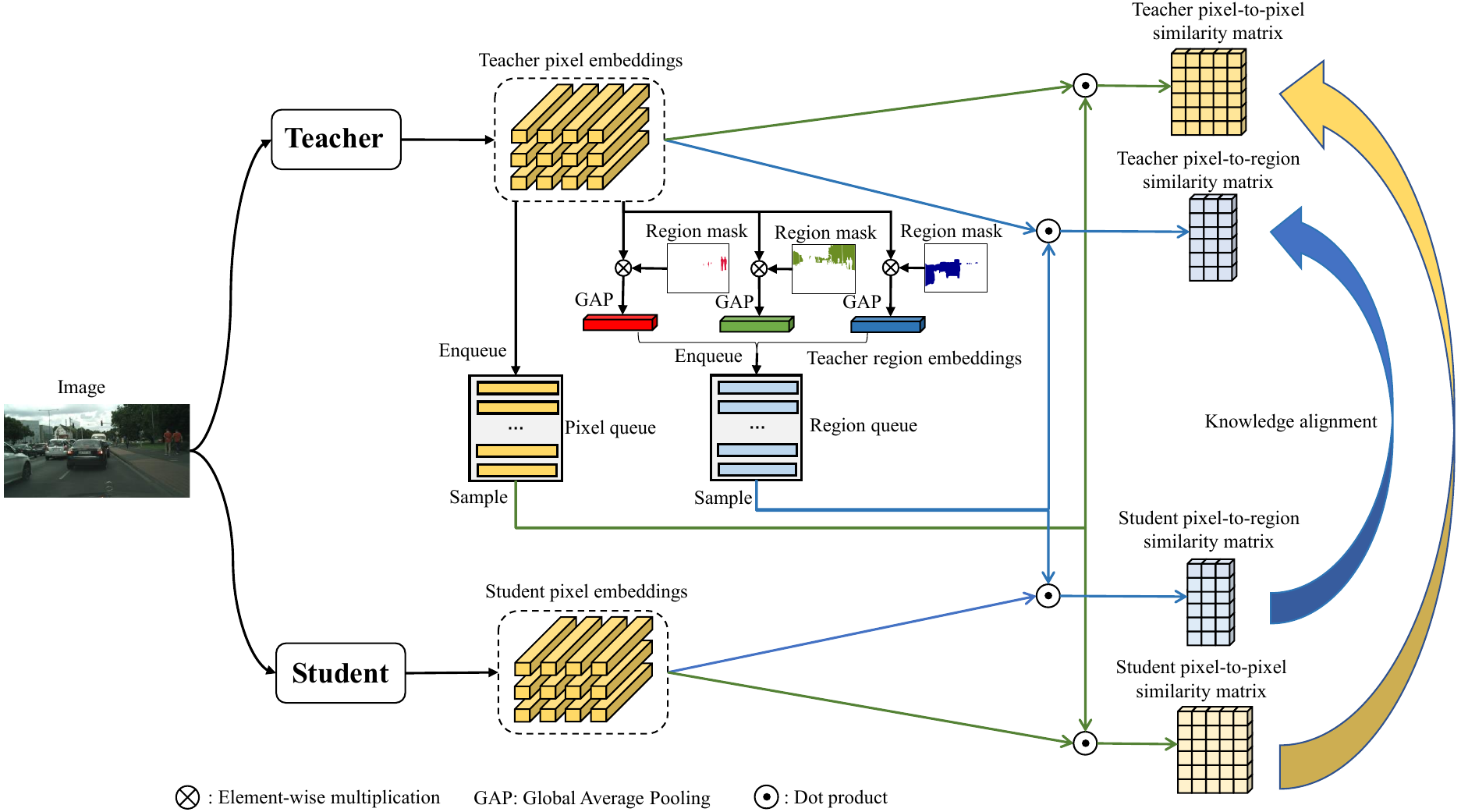}%
			\label{mem_kd}}
		\caption{Overview of our proposed pixel-to-pixel and pixel-to-region distillation.}
		\label{mini_mem_kd}
	\end{figure*}

	We show the illustration of mini-batch-based pixel-to-pixel distillation in Figure~\ref{mini_pixel_kd}. Given a mini-batch $\{\bm{x}_{n}\}_{n=1}^{N}$, the segmentation network extracts $N$ structured feature maps $\{\mathbf{F}_{n}\in \mathbb{R}^{H\times W\times D}\}_{n=1}^{N}$ from $N$ input images. We preprocess each pixel embedding in $\mathbf{F}_{n}$ by $l_{2}$-normalization. For easy notation, we reshape the spatial dimension of $\{\mathbf{F}_{n}\in \mathbb{R}^{H\times W\times D}\}_{n=1}^{N}$ to $\{\mathbf{F}_{n}\in \mathbb{R}^{A\times D}\}_{n=1}^{N}$, where $A=H\times W$. For the $i$-th image $\bm{x}_{i}$ and the $j$-th image $\bm{x}_{j}$, $i,j\in \{1,2,\cdots,N\}$, we can calculate the cross-image pair-wise similarity matrix $\mathbf{S}_{ij}=\mathbf{F}_{i}\mathbf{F}_{j}^\top\in \mathbb{R}^{A\times A}$.
	The relational matrix $\mathbf{S}_{ij}$ captures the cross-image pair-wise correlations among pixels. 
	
	We guide the pair-wise similarity matrix of $\mathbf{S}^{s}_{ij}$ produced from the student to align that of $\mathbf{S}^{t}_{ij}$ produced from the teacher. The distillation process is formulated as:
	\begin{equation}
		L_{p2p}(\mathbf{S}^{s}_{ij},\mathbf{S}^{t}_{ij})=\frac{1}{A}\sum_{a=1}^{A}KL(\sigma(\frac{\mathbf{S}^{s}_{ij|a,:}}{\tau})||\sigma(\frac{\mathbf{S}^{t}_{ij|a,:}}{\tau})).
	\end{equation}
	Here, $\mathbf{S}_{ij|a,:}$ denotes the $a$-th row vector of $\mathbf{S}_{ij}$. We normalize each row similarity distribution of $\mathbf{S}_{ij}$ to a probability distribution with a temperature $\tau$ by softmax function $\sigma$. The magnitude gaps would be removed between the student and teacher networks due to the softmax normalization. $KL$ is used to align each row-wise probability distribution. We perform pixel-to-pixel distillation every two of $N$ images:
	\begin{equation}
		L_{batch\_p2p}=\frac{1}{N^{2}}\sum_{i=1}^{N}\sum_{j=1}^{N}L_{p2p}(\mathbf{S}^{s}_{ij},\mathbf{S}^{t}_{ij}).
	\end{equation}

	\subsubsection{Memory-based Pixel-to-Pixel Distillation}
	Although mini-batch-based distillation could capture cross-image relations to some extent, it is difficult to model dependencies among pixels from global images, since the batch size per GPU of segmentation task is often small, \emph{e.g.} 1 or 2. To address this problem, we introduce an online pixel queue that can store massive pixel embeddings in the memory bank generated from the past mini-batches. It allows us to retrieve abundant embeddings efficiently. The usage of memory bank dates back to self-supervised learning~\cite{wu2018unsupervised}. This is because a large number of negative samples are pivotal for unsupervised contrastive learning, and the mini-batch size limits available contrastive samples. 
	
	In the context of the dense segmentation task, each image would contain a vast number of pixel samples, and most pixels in the same object region are often homogeneous. Therefore, storing all pixel embeddings may learn redundant relational knowledge and slow down the distillation process. Moreover, saving several last batches to the queue may also damage the diversity of pixel embeddings. Thus we maintain a class-aware pixel queue $\bm{Q}_{p}\in \mathbb{R}^{C\times N_{p}\times D}$, where $N_{p}$ is the number of pixel embeddings per class and $D$ is the embedding size. For each image in the mini-batch, we only randomly sample a small number, \emph{i.e.} $V$ ($V\ll N_{p}$), of pixel embeddings from the same class and push them into the pixel queue $\bm{Q}_{p}$. The queue is progressively
	updated under the “first-in-first-out” strategy as distillation proceeds.
	
	Inspired by Fang \textit{et al.}~\cite{fang2021seed}, we adopt a shared pixel queue between the student and teacher networks and store pixel embeddings generated from the teacher during the distillation process. Given an input image $\bm{x}_{n}$, the generated pixel embeddings of the student and teacher networks are $\mathbf{F}_{n}^{s}\in \mathbb{R}^{A\times D}$ and $\mathbf{F}_{n}^{t}\in \mathbb{R}^{A\times D}$, respectively. Each pixel embedding of $\mathbf{F}_{n}^{s}$ and $\mathbf{F}_{n}^{t}$ is preprocessed by $l_{2}$-normalization. We regard $\mathbf{F}_{n}^{s}$ and $\mathbf{F}_{n}^{t}$ as anchors and sample $K_{p}$ contrastive embeddings $\{\bm{v}_{k}\in \mathbb{R}^{D}\}_{k=1}^{K_{p}}$ randomly from the pixel queue $\bm{Q}_{p}$. Here, we adopt a class-balanced sampling since the numbers of pixels from various classes often conform to a long-tailed distribution. For easy notation, $\mathbf{V}_{p}=[\bm{v}_{1},\bm{v}_{2},\cdots,\bm{v}_{K_{p}}]\in \mathbb{R}^{K_{p}\times D}$ is a concatenation of $\{\bm{v}_{k}\}_{k=1}^{K_{p}}$ along the row dimension.  Then we model the pixel similarity matrix between the anchors and contrastive embeddings for the student and teacher as $\mathbf{P}^{s}$ and $\mathbf{P}^{t}$:
	\begin{equation}
		\mathbf{P}^{s}=\mathbf{F}_{n}^{s}\mathbf{V}_{p}^{\top}\in \mathbb{R}^{A\times K_{p}},\  \mathbf{P}^{t}=\mathbf{F}_{n}^{t}\mathbf{V}_{p}^{\top}\in \mathbb{R}^{A\times K_{p}}.
	\end{equation}

	The teacher network often shows a better pixel similarity matrix than the student. We force the student's $\mathbf{P}^{s}$ to mimic the teacher's $\mathbf{P}^{t}$ for penalizing the difference. Similar to Section~\ref{minibatch}, we apply softmax normalization on each row distribution of $\mathbf{P}^{s}$ and $\mathbf{P}^{t}$ and perform pixel-to-pixel distillation via KL-divergence loss. It is formulated as follows:
	\begin{equation}
		L_{memory\_p2p}=\frac{1}{A}\sum_{a=1}^{A}KL(\sigma(\frac{\mathbf{P}^{s}_{a,:}}{\tau})||\sigma(\frac{\mathbf{P}^{t}_{a,:}}{\tau})).
		\label{memory_p2p}
	\end{equation}
	
	After each iteration, we push $V_p$ teacher pixel embeddings per class into the pixel queue $\bm{Q}_{p}$. Because the teacher is pre-trained and frozen, it can provide consistent feature embeddings during the distillation process. Therefore, we can naturally avoid the inconsistent problem between the anchor and dequeued features appeared in previous contrastive learning~\cite{he2020momentum,hu2021region,wang2021exploring}.

	\subsubsection{Memory-based Pixel-to-Region Distillation}
	Discrete pixel embeddings may not fully capture image content. Therefore, we introduce an online region queue that can store massive more representative region embeddings in the memory bank.  Beyond pixel-to-pixel distillation, we further construct pixel-to-region distillation to model the relations between pixels and class-wise region embeddings across global images. Each region embedding represents the feature center of one semantic class in an image. We formulate the region embedding of a class by averagely pooling all the pixel embeddings belonging to that class in a single image. 
	
	We maintain a region queue $\bm{Q}_{r}\in \mathbb{R}^{C\times N_{r}\times D}$ during the distillation process, where $N_{r}$ is the number of region embeddings per class and $D$ is the embedding size. For each iteration, we sample $K_{r}$ contrastive region embeddings $\{\bm{r}_{k}\in \mathbb{R}^{D}\}_{k=1}^{K_{r}}$ from $\bm{Q}_{r}$ in a class-balanced manner. For easy notation, $\mathbf{V}_{r}=[\bm{r}_{1},\bm{r}_{2},\cdots,\bm{r}_{K_{r}}]\in \mathbb{R}^{K_{r}\times D}$ is a concatenation of $\{\bm{r}_{k}\}_{k=1}^{K_{r}}$ along the row dimension. Given an input image $\bm{x}_{n}$, we model the pixel-to-region similarity matrix from $\mathbf{F}_{n}^{s}\in \mathbb{R}^{A\times D}$ and $\mathbf{F}_{n}^{t}\in \mathbb{R}^{A\times D}$ to region embeddings $\mathbf{V}_{r}$ as $\mathbf{R}^{s}$ and $\mathbf{R}^{t}$:
	\begin{equation}
		\mathbf{R}^{s}=\mathbf{F}_{n}^{s}\mathbf{V}_{r}^{\top}\in \mathbb{R}^{A\times K_{r}},\  \mathbf{R}^{t}=\mathbf{F}_{n}^{t}\mathbf{V}_{r}^{\top}\in \mathbb{R}^{A\times K_{r}}.
	\end{equation}
	
	Similar to the Equ.~(\ref{memory_p2p}), we distill normalized pixel-to-region similarity matrix between the student and teacher networks via KL-divergence loss:
	\begin{equation}
		L_{memory\_p2r}=\frac{1}{A}\sum_{a=1}^{A}KL(\sigma(\frac{\mathbf{R}^{s}_{a,:}}{\tau})||\sigma(\frac{\mathbf{R}^{t}_{a,:}}{\tau})).
		\label{memory_p2r}
	\end{equation}
	For each mini-batch, we push all teacher region embeddings into the region queue $\bm{Q}_{r}$. The overview of our proposed memory-based distillation is shown in Figure~\ref{mem_kd}.

	\subsection{Channel-wise Cross-Image Relational Knowledge Distillation}
	\begin{figure}[t]  
		\centering 
		\includegraphics[width=1\linewidth]{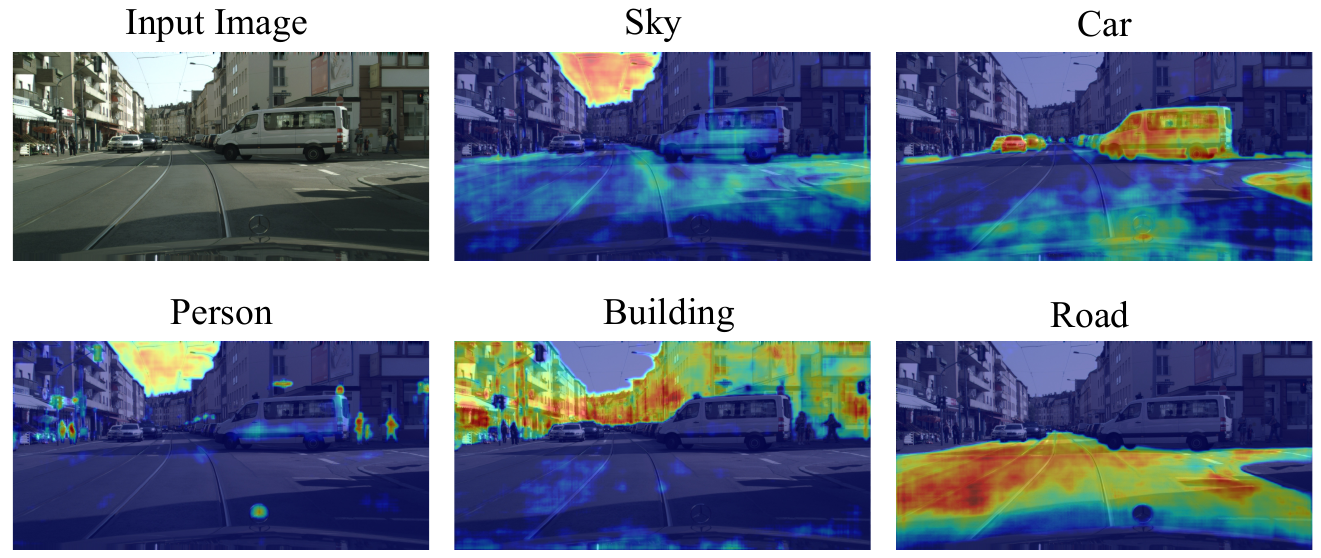}
		\caption{Visualization of channel semantics over DeepLabV3 on Cityscapes. Each heatmap corresponds to one channel.} 
		\label{channel_semantic}
	\end{figure}
	As shown in Figure~\ref{channel_semantic}, we found that various channels in a semantic segmentation network often encode different category semantics. This observation is also verified by Shu \textit{et al.}~\cite{shu2021channel}. This property motivates us to model cross-image relational knowledge from channel interaction.

	\begin{figure*}[!t]
		\centering
		\subfloat[Mini-batch channel distillation]{\includegraphics[width=0.32\textwidth]{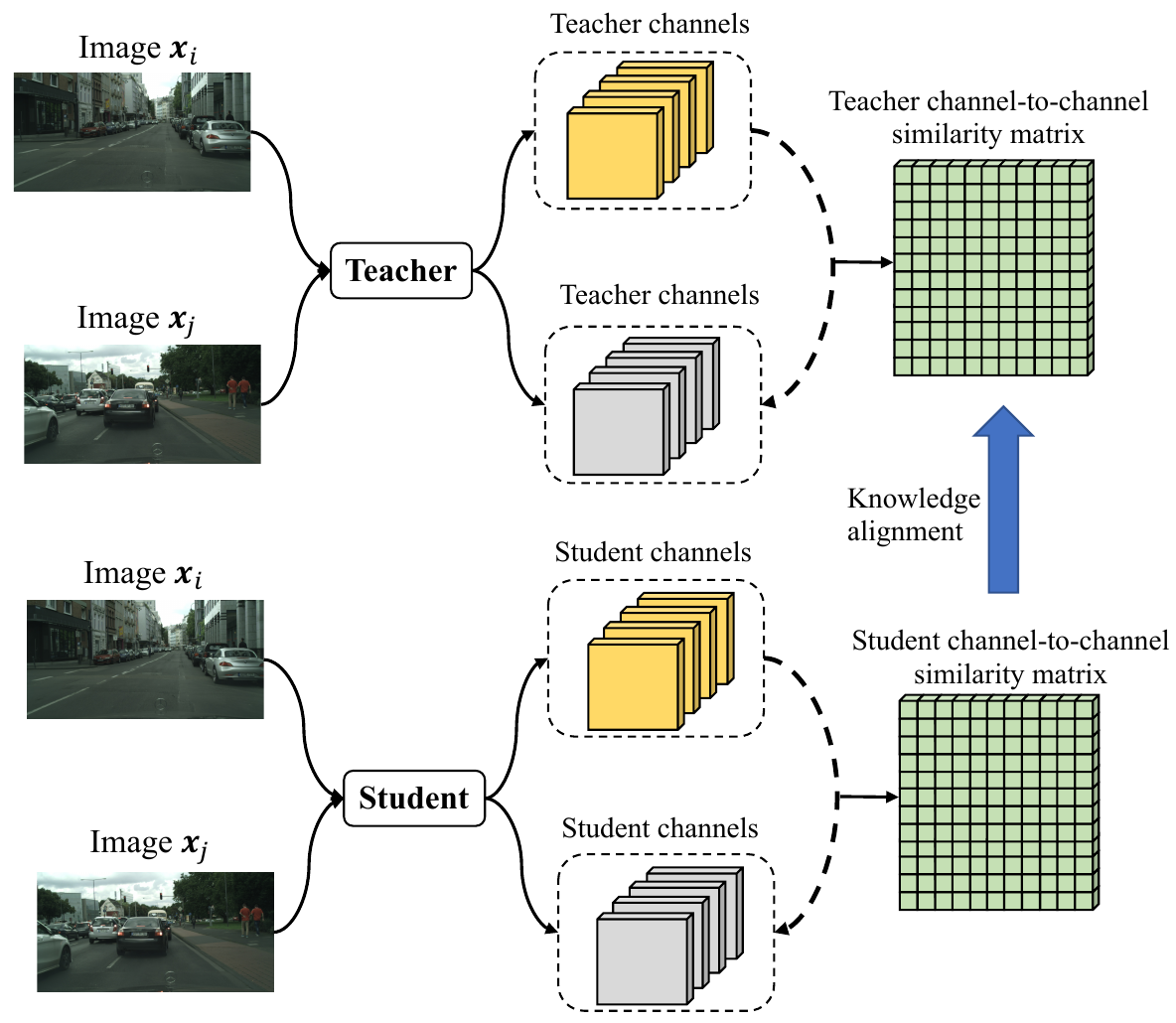}%
			\label{mini_batch_channel}}
		\subfloat[Memory-based channel distillation]{\includegraphics[width=0.35\textwidth]{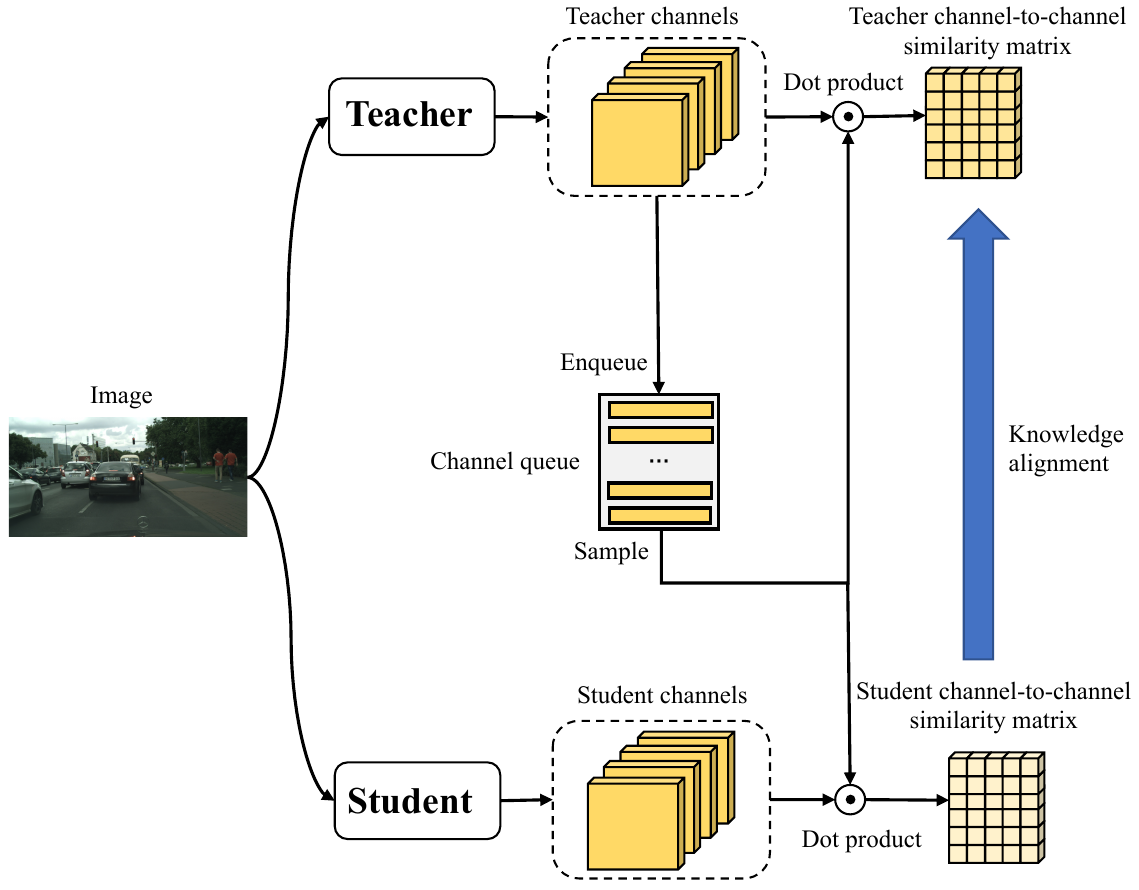}%
			\label{memory_channel}}
		\subfloat[Contextual feature distillation]{\includegraphics[width=0.3\textwidth]{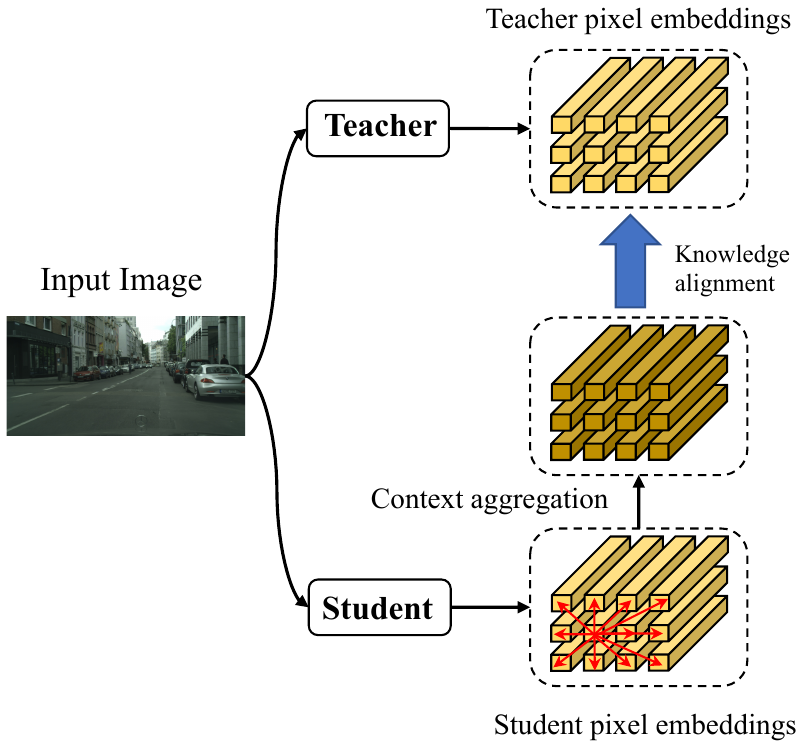}%
			\label{fd}}
		\caption{Overview of our proposed channel and feature distillation.}
		\label{channel_feature}
	\end{figure*}
	
	\subsubsection{Mini-batch-based channel-to-channel distillation}
	
	The illustration of mini-batch-based channel-to-channel distillation is shown in Figure~\ref{mini_batch_channel}. Given the $N$ structured feature maps $\{\mathbf{F}_{n}\in \mathbb{R}^{H\times W\times D}\}_{n=1}^{N}$ from $N$ input images from the mini-batch $\{\bm{x}_{n}\}_{n=1}^{N}$, we preprocess each channel in $\mathbf{F}_{n}$ by $l_{2}$-normalization. For easy notation, we flatten the spatial dimension to $\{\mathbf{F}_{n}\in \mathbb{R}^{A\times D}\}_{n=1}^{N}$ and then transpose, deriving the set of channel embeddings $\{\mathbf{F}_{n}^{\top}\in \mathbb{R}^{D\times A}\}_{n=1}^{N}$, where $A=H\times W$. In pixel-to-pixel distillation, the pixels from the same spatial location between teacher and student often represent similar semantics. Unlike spatial dimension, the teacher and student with independent training often show different semantics in the same channel location. To promote channel-to-channel distillation, the same channel location should have similar semantics between teacher and student as the prerequisite. To align channel semantics between teacher and student, we first apply a Mean Squared Error (MSE) loss for each channel:
	\begin{equation}
		L_{channel\_fd}=\frac{1}{D}\sum_{d=1}^{D}\left \| \mathbf{F}_{n|:,d}^s-\mathbf{F}_{n|:,d}^t \right \|  ^{2}_F.
		\label{channel_fd_equ}
	\end{equation}
	In this way, the student could learn similar channel distributions with the teacher. When the student and teacher have different feature dimensions, we first attach a lightweight projection head to the student feature map. The projection head consists of two 1×1 convolutional layers with batch normalization and ReLU. It maps the student feature dimension to the teacher feature dimension. Eq. (\ref{channel_fd_equ}) is then applied after this projection. The projection head is only used during training and is discarded during inference.
		We would like to clarify that we do not assume that the original student channels and teacher channels are naturally aligned. Instead, the projection head learns a teacher-compatible channel space under the supervision of channel feature alignment and relational distillation. The one-to-one alignment is therefore performed in \textbf{the projected space}, rather than between the original raw channels of two heterogeneous networks.

	For the $i$-th image $\bm{x}_{i}$ and the $j$-th image $\bm{x}_{j}$, $i,j\in \{1,2,\cdots,N\}$, we can calculate the cross-image pair-wise channel similarity matrix $\mathbf{G}_{ij}=\mathbf{F}_{i}^\top\mathbf{F}_{j}\in \mathbb{R}^{D\times D}$.
	The channel relational matrix $\mathbf{G}_{ij}$ captures the cross-image pair-wise correlations among semantic channels.  
	
	We guide the pair-wise channel similarity matrix of $\mathbf{G}^{s}_{ij}$ produced from the student to align that of $\mathbf{G}^{t}_{ij}$ produced from the teacher. The distillation process is formulated as:
	\begin{equation}
		L_{c2c}(\mathbf{G}^{s}_{ij},\mathbf{G}^{t}_{ij})=\frac{1}{D}\sum_{d=1}^{D}KL(\sigma(\frac{\mathbf{G}^{s}_{ij|d,:}}{\tau})||\sigma(\frac{\mathbf{G}^{t}_{ij|d,:}}{\tau})).
	\end{equation}
	Here, $\mathbf{G}_{ij|d,:}$ denotes the $d$-th row vector of $\mathbf{G}_{ij}$. Similar to pixel-to-pixel distillation, we normalize each row similarity distribution of $\mathbf{G}_{ij}$ to a probability distribution with a temperature $\tau$ by softmax function $\sigma$. $KL$ is used to align each row-wise probability distribution. We perform channel-to-channel distillation every two of $N$ images:
	\begin{equation}
		L_{batch\_c2c}=\frac{1}{N^{2}}\sum_{i=1}^{N}\sum_{j=1}^{N}L_{c2c}(\mathbf{G}^{s}_{ij},\mathbf{G}^{t}_{ij}).
	\end{equation}

	\subsubsection{Memory-based channel-to-channel distillation}
	As discussed above, mini-batch-based distillation could not capture cross-image relations from global images. To address this problem, we introduce an online channel queue $\bm{Q}_{c}\in \mathbb{R}^{N_{c}\times A}$ that can store massive channel embeddings in the memory bank generated from the past mini-batches, where $N_{c}$ is the number of channels and $A$ is the embedding size. It allows us to retrieve abundant channel embeddings efficiently.

	For each iteration, we randomly sample $K_c$ contrastive channel embeddings $\{\bm{c}_{k}\in \mathbb{R}^{A}\}_{k=1}^{K_{c}}$ from $\bm{Q}_{c}$. For easy notation, $\mathbf{V}_{c}=[\bm{c}_{1},\bm{c}_{2},\cdots,\bm{c}_{K_{c}}]\in \mathbb{R}^{K_{c}\times A}$ is a concatenation of $\{\bm{c}_{k}\}_{k=1}^{K_{c}}$ along the row dimension. Given an input image $\bm{x}_{n}$, we model the channel-to-channel similarity matrix from $(\mathbf{F}_{n}^{s})^\top\in \mathbb{R}^{D\times A}$ and $(\mathbf{F}_{n}^{t})^\top\in \mathbb{R}^{D\times A}$ to channel embeddings $\mathbf{V}_{c}$ as $\mathbf{C}^{s}$ and $\mathbf{C}^{t}$:
	\begin{equation}
		\mathbf{C}^{s}=(\mathbf{F}_{n}^{s})^\top\mathbf{V}_{c}^{\top}\in \mathbb{R}^{D\times K_{c}},\  \mathbf{C}^{t}=(\mathbf{F}_{n}^{t})^\top\mathbf{V}_{c}^{\top}\in \mathbb{R}^{D\times K_{c}}.
	\end{equation}
	
	Similar to the Equ.~(\ref{memory_p2p}), we distill normalized channel-to-channel similarity matrix between the student and teacher networks via KL-divergence loss:
	\begin{equation}
		L_{memory\_c2c}=\frac{1}{D}\sum_{d=1}^{D}KL(\sigma(\frac{\mathbf{C}^{s}_{d,:}}{\tau})||\sigma(\frac{\mathbf{C}^{t}_{d,:}}{\tau})).
		\label{memory_p2r}
	\end{equation}
	After each iteration, we push $V_c$ teacher channel embeddings into the channel queue $\bm{Q}_{c}$.
	
	\subsection{Feature Distillation with Context Aggregation}

	Cross-image relational modeling often captures structured knowledge. Beyond relation distillation, feature distillation~\cite{romero2014fitnets,yang2024clip} has been demonstrated to reduce the performance gap between teacher and student. We further explore feature distillation of pixel embeddings to increase the similarity of pixel-wise representations between teacher and student. The traditional feature distillation methods often directly align pixel-wise embeddings from the original teacher and student feature maps. However, the pixels over the same spatial location from teacher and student may not have consistent semantics, since the teacher is often equipped with a larger kernel size than the student,  which means that the teacher pixel extracts semantic information from a wider receptive field.
	
	To mitigate the receptive field gap between teacher and student, we utilize context aggregation over the student feature map to augment student pixel-wise representations using wider contextual semantic information. Given the feature map $\mathbf{F}\in \mathbb{R}^{A\times D}$, where $A$ is the number of pixels and $D$ is the pixel embedding dimension, we linearly transform it to $\bm{Q}$, $\bm{K}$, and $\bm{V}$ as the concepts of query, key and value. We use the self-attention mechanism~\cite{vaswani2017attention} to model context aggregation for pixel representations:
	\begin{equation}
		\hat{\mathbf{F}}={\rm softmax}(\frac{\bm{Q}\bm{K}^\top}{\sqrt{D}})\bm{V}.
	\end{equation}
	We concatenate the original pixel representations $\mathbf{F}$ and pixel contextual representations $\hat{\mathbf{F}}$ with transformations as the final augmented pixel representations $\bar{\mathbf{F}}$:
	\begin{equation}
		\bar{\mathbf{F}}=\mathcal{T}_2(\mathbf{F} || \mathcal{T}_1(\hat{\mathbf{F}})).
	\end{equation}
	Here, $||$ denotes the concatenation operator. $\mathcal{T}_1$ and $\mathcal{T}_2$ represent a sequence of transformation functions: $1\times 1$ convolution$\to $batch normalization$\to $ReLU.
	
	We further conduct feature distillation by guiding the augmented pixel representations $\bar{\mathbf{F}}^s$ from the student to mimic the teacher pixel representations $\mathbf{F}^t$ by MSE:
	\begin{equation}
		L_{context\_fd}=\frac{1}{AD}\left \| \bar{\mathbf{F}}^s- \mathbf{F}^t\right \|  ^{2}_F.
	\end{equation}
	Benefiting from the contextual feature distillation, the student would learn better pixel  representations with meaningful context semantics from the teacher, leading to improved semantic segmentation capability.

	The context aggregation module can indeed be viewed as an auxiliary distillation head during training. However, it is not used to enhance the inference-time capacity of the student. Instead, it provides a teacher-compatible feature transformation for the distillation objective.
	It is a training-only semantic adapter that improves the quality of feature distillation without introducing any inference overhead. This is consistent with many distillation methods that use auxiliary heads or projection layers during training and remove them during inference.

	\subsection{Overall Framework}
	We summarize our mini-batch-based and memory-based relational distillation, as well as feature distillation together to train the student network. We also employ the conventional pixel-wise cross-entropy task loss $L_{task}$ (Equ.~(\ref{task})) and class probability distillation loss $L_{kd}$ (Equ.~(\ref{kd})) as the basic losses. The overall loss is formulated as:
	
	\begin{align}
		&L_{CIRKDV2}=L_{task}+\underbrace{L_{kd}}_{probability\ distillation} \notag \\
		&+\underbrace{\alpha(L_{batch\_p2p}+L_{batch\_c2c})}_{mini-batch-based\ relational\ distillation} \notag \\
		& +\underbrace{\beta( L_{memory\_p2p}+L_{memory\_p2r}+L_{memory\_c2c})}_{memory-based\ relational\ distillation} \notag \\
		& +\underbrace{\gamma L_{channel\_fd}
			+\lambda L_{context\_fd}}_{feature\ distillation}.
		\label{overall_loss}
	\end{align}
	Here, $\alpha$, $\beta$, $\gamma$, and $\lambda$ are weight coefficients of mini-batch-based relational distillation, memory-based relational distillation, channel feature distillation, and contextual feature distillation, respectively. We adopt the same weight $\alpha$ for mini-batch-based $L_{batch\_p2p}$ as well as $L_{batch\_c2c}$, and the same weight $\beta$ for  $L_{memory\_p2p}$, $L_{memory\_p2r}$, and $L_{memory\_c2c}$, because losses belonging to the identical type have the consistent order of magnitudes. We set $\alpha=1$, $\beta=0.1$, $\gamma=100$, and $\lambda=10$. Empirically, we find our CIRKDV2 is not sensitive to coefficients. When the student and teacher networks mismatch the embedding size, we attach a projection head to the student network.  It can map the student's embeddings to match the teacher's dimension. The projection head is composed of two $1\times 1$ convolutional layers with batch normalization and ReLU. It would be discarded at the inference phase without introducing extra costs. In Algorithm \ref{cirkd_alg}, we use pseudo-code to illustrate the overall training pipeline of CIRKDV2.
	
	\begin{algorithm}[tb]
		\caption{Cross-Image Relational KD (CIRKDV2)}
		\begin{algorithmic}[1] 
			\label{cirkd_alg}
			\STATE
			Initialize the pixel queue $\bm{Q}_{p}$, region queue $\bm{Q}_{r}$, and channel queue $\bm{Q}_{c}$  with random unit vectors. 
			\WHILE{the student network has not converged}
			\STATE 	Sample a mini-batch.
			\STATE Generate the student and teacher feature maps.
			\STATE  Compute the mini-batch-based pixel-to-pixel distillation loss $L_{batch\_p2p}$ and channel-to-channel distillation loss $L_{batch\_c2c}$.
			\STATE Compute channel alignment loss $L_{channel\_fd}$.
			\STATE 	Sample contrastive pixel, region, and channel embeddings from $\bm{Q}_{p}$, $\bm{Q}_{r}$, and $\bm{Q}_{c}$.
			\STATE Compute the memory-based pixel-to-pixel loss $L_{memory\_p2p}$, pixel-to-region loss $L_{memory\_p2r}$, and channel-to-channel loss $L_{memory\_c2c}$.
			\STATE Compute contextual feature distillation loss $L_{context\_fd}$.
			\STATE Update the student w.r.t the overall loss $L_{CIRKDV2}$.
			\STATE Enqueue the current teacher pixel, region, and channel embeddings to $\bm{Q}_{p}$, $\bm{Q}_{r}$, and $\bm{Q}_{c}$.
			\STATE Dequeue the earliest pixel, region, and channel embeddings from $\bm{Q}_{p}$, $\bm{Q}_{r}$, and $\bm{Q}_{c}$.
			\ENDWHILE
		\end{algorithmic}
		
	\end{algorithm}

	\section{Main Experiments}
	\subsection{Experimental Setup}
	\textbf{Datasets.} We employ four popular semantic segmentation datasets to conduct our experiments. 
	\begin{enumerate}
		\item \textbf{Cityscapes}~\cite{cordts2016cityscapes} is an urban scene parsing dataset that contains 5000 finely annotated images, where 2,975/500/1525 images are used for \texttt{train/val/test}. The segmentation performance is reported on 19 classes.
		\item \textbf{ADE20K}~\cite{zhou2017scene}  contains 20K/2K/3K images for \texttt{train/val/test} with 150 semantic categories under diverse scenes.
		\item \textbf{Pascal VOC}~\cite{everingham2010pascal} is a visual object segmentation dataset that includes 20 foreground object categories and one background class. We adopt the augmented data with extra annotations provided by Hariharan \textit{et al.}~\cite{hariharan2011semantic}. The resulting dataset contains 10,582/1,449/1,456 images for \texttt{train/val/test}. 
		\item \textbf{COCO-Stuff-164K}~\cite{caesar2018coco}  contains 164K images from COCO-2017~\cite{lin2014microsoft}: 118K for
		training, 5K for validation, 20K for test-dev, and 20K for the test-challenge. It covers 172 classes: 80 thing classes, 91 stuff classes, and 1 class “unlabeled”.
		
	\end{enumerate}

	\textbf{Evaluation metrics.} Following the standard setup, we employ mean
	Intersection-over-Union (mIoU) to measure the segmentation performance. The model size is measured by the number of network parameters, and the computing complexity is represented by the sum of floating point operations (FLOPs) in one forward step using a fixed input size, where the input size is 1024$\times$2048 for Cityscapes, and 512$\times$512 for other datasets. 
	\begin{figure*}[tbp]  
		\centering 
		\includegraphics[width=1\linewidth]{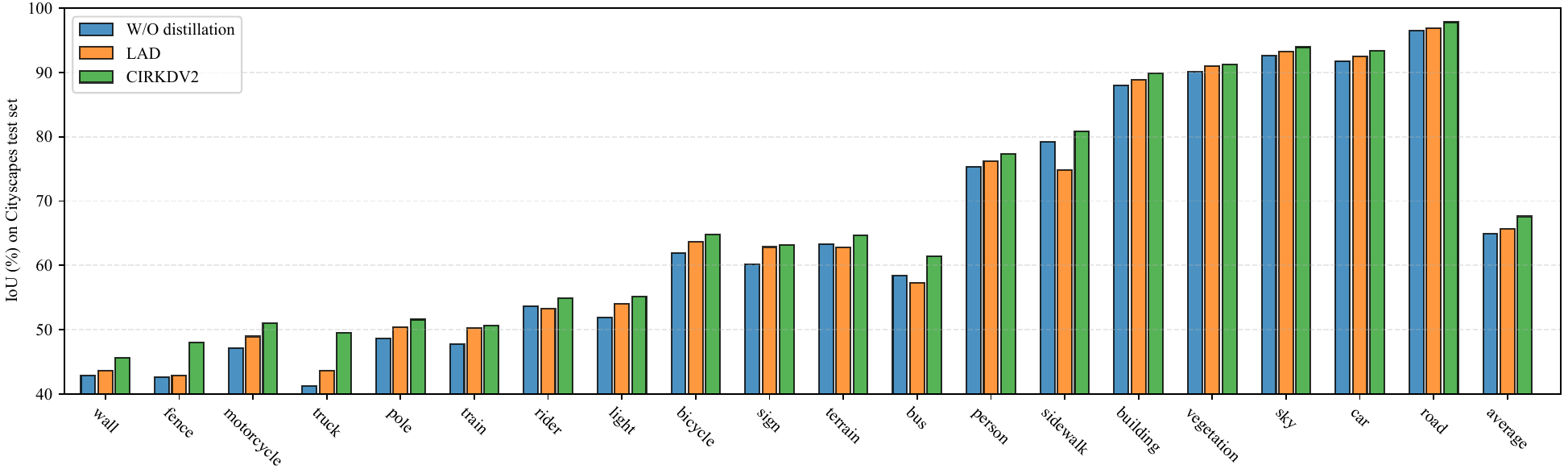}
		\caption{Illustration of individual class IoU scores over the student network DeepLabV3-MobileNetV3-Small with baseline (W/O distillation), the state-of-the-art LAD and our proposed CIRKDV2 on Cityscapes test set. Our CIRKDV2 can consistently improve individual class IoU scores compared to the baseline and LAD, especially for those challenging classes with low IoU scores. }
		\label{cityscape_iou}
		\vspace{-0.2cm}
	\end{figure*}

	\textbf{Network architectures.} Besides SegFormer~\cite{xie2021segformer}, we use the segmentation framework DeepLabV3~\cite{chen2018encoder} with ResNet-101 backbone~\cite{he2016deep} as the powerful teacher network. For student networks, we use diverse segmentation architectures to verify the effectiveness of distillation methods. Specifically, we adopt various combinations of DeepLabV3, PSPNet~\cite{zhao2017pyramid}, and UperNet~\cite{xiao2018unified} with different light-weight backbones of ResNet-18~\cite{he2016deep}, MobileNetV3~\cite{howard2019searching}, and MobileViT~\cite{mehta2021mobilevit}. For SegFormer~\cite{xie2021segformer} distillation, the teacher and student backbones are MiT-B4 and MiT-B0, respectively. All backbones are initialized by pretrained weights from ImageNet-1K~\cite{russakovsky2015imagenet}.
	
	\textbf{Training details.} Following the standard data augmentation, we employ random flipping and scaling in the range of $[0.5,2]$. For Convolutional Neural Network (CNN)-based networks, we use SGD optimizer with weight decay 0.0001. For Cityscapes, the batch size is 8, and the initial learning rate is 0.01. For other datasets, the batch size is 16, and the initial learning rate is 0.02. The number of the total training iterations is 80K. The learning rate is decayed by $(1-\frac{iter}{total\_iter})^{0.9}$ following the polynomial annealing policy~\cite{chen2017rethinking}. For MobileViT, we follow the official setup~\cite{mehta2021mobilevit} by adopting AdamW optimizer with an initial learning rate 0.0009 and weight decay 0.01. Following the official setup~\cite{xie2021segformer}, SegFormer is optimized by AdamW with an initial learning rate 0.0002 and weight decay 0.0001. For crop size during the training phase, we use 1024$\times$1024 on SegFormer and 512$\times$1024 on other networks for Cityscapes. Other datasets follow the 512$\times$512 crop size.

	\textbf{Evaluation details.} Following the common protocol with previous works~\cite{liu2019structured,yang2022cross}, we evaluate the segmentation performance under a single-scale setting over the original image size. Moreover, we report test mIoU for Cityscapes by submitting results to the official Cityscapes test server\footnote{https://www.cityscapes-dataset.com/submit/}. Beyond Cityscapes, we report validation mIoU for other semantic segmentation datasets.
	
	\textbf{Compared distillation methods.} We compare our proposed CIRKDV2 with state-of-the-art segmentation distillation methods: SKD~\cite{liu2019structured}, IFVD~\cite{wang2020intra}, CWD~\cite{shu2021channel}, DSD~\cite{feng2021double}, CIRKD~\cite{yang2022cross}, APD~\cite{tian2022adaptive}, Af-DCD~\cite{fan2023augmentation}, and LAD~\cite{liu2024rethinking}. We re-run all methods using author-provided code under the same pre-trained teacher.

	\begin{table}[tbp]
		\centering
		\caption{Performance comparison with the state-of-the-art distillation methods over various CNN-based segmentation networks on Cityscapes.}
		\resizebox{1.\linewidth}{!}{
			\begin{tabular}{l|c|c|cc}  
				\hline
				Method&Params&FLOPs& mIoU (\%) \\ 
				\hline
				T: DeepLabV3-ResNet-101 &79.1M &3228.5G &78.30   \\
				\hline
				S: DeepLabV3-ResNet-18 &\multirow{10}{*}{22.8M} &\multirow{10}{*}{972.4G} &  73.56 \\
				+SKD~\cite{liu2019structured} & & & 74.21   \\
				+IFVD~\cite{wang2020intra} & & &  74.35  \\
				+CWD~\cite{shu2021channel} & & &  74.28  \\
				+DSD~\cite{feng2021double} & & &  74.05  \\
				+CIRKD~\cite{yang2022cross} & & &  75.15  \\
				+APD~\cite{tian2022adaptive}& & & 74.68 \\
				+Af-DCD~\cite{fan2023augmentation}& &  & 75.22 \\
				+LAD~\cite{liu2024rethinking} & & &  \underline{75.28}  \\
				+CIRKDV2 (Ours) & & &  \textbf{75.60}  \\
				\hline
				S: UperNet-ResNet-18 &\multirow{10}{*}{16.6M} &\multirow{10}{*}{650.7G} & 68.90   \\
				+SKD~\cite{liu2019structured} & & &  70.34  \\
				+IFVD~\cite{wang2020intra} & & & 69.84   \\
				+CWD~\cite{shu2021channel} & & & 71.13   \\
				+DSD~\cite{feng2021double} & & & 70.84   \\
				+CIRKD~\cite{yang2022cross} & & & 71.46  \\
				+APD~\cite{tian2022adaptive}& & &  70.84 \\
				+Af-DCD~\cite{fan2023augmentation}& &  & 70.82 \\
				+LAD~\cite{liu2024rethinking} & & &  \underline{71.57}  \\
				+CIRKDV2 (Ours) & & &  \textbf{72.11}  \\
				\hline
				S: DeepLabV3-MobileNetV3-Small &\multirow{10}{*}{3.3M} &\multirow{10}{*}{130.2G} & 65.05  \\
				+SKD~\cite{liu2019structured} & & & 66.14   \\
				+IFVD~\cite{wang2020intra} & & & 65.46   \\
				+CWD~\cite{shu2021channel} & & & 66.73   \\
				+DSD~\cite{feng2021double} & & &  66.32  \\
				+CIRKD~\cite{yang2022cross} & & &  \underline{66.84} \\
				+APD~\cite{tian2022adaptive}& & &  65.72\\
				+Af-DCD~\cite{fan2023augmentation}& &  & 66.53 \\
				+LAD~\cite{liu2024rethinking} & & &  65.65  \\
				+CIRKDV2 (Ours) & & &   \textbf{67.62} \\
				\hline
				S: PSPNet-MobileNetV3-Small &\multirow{10}{*}{2.6M} &\multirow{10}{*}{58.4G} &  62.78   \\
				+SKD~\cite{liu2019structured} & & &   63.53 \\
				+IFVD~\cite{wang2020intra} & & &  63.24  \\
				+CWD~\cite{shu2021channel} & & &  63.64  \\
				+DSD~\cite{feng2021double} & & &  64.14   \\
				+CIRKD~\cite{yang2022cross} & & &  64.52 \\
				+APD~\cite{tian2022adaptive}& & & 64.23 \\
				+Af-DCD~\cite{fan2023augmentation}& &  &  \underline{64.68}\\
				+LAD~\cite{liu2024rethinking} & & &   64.32 \\
				+CIRKDV2 (Ours) & & &   \textbf{65.42}  \\
				\hline
				
		\end{tabular}}
		\vspace{-0.1cm}
		
		\label{citys} 
		\vspace{-0.5cm}
	\end{table}

	\begin{table}[tbp]
		\centering
		\caption{Performance comparison with the state-of-the-art distillation methods over various MobileViT-based segmentation networks on Cityscapes.}
		\resizebox{1.\linewidth}{!}{
			\begin{tabular}{l|c|c|cc}  
				\hline
				Method&Params&FLOPs& mIoU (\%) \\ 
				\hline
				T: DeepLabV3-ResNet-101 &79.1M &3228.5G &78.30   \\
				\hline
				S: DeepLabV3-MobileViT-XXS &\multirow{10}{*}{2.4M} &\multirow{10}{*}{473.6G} &66.24    \\
				+SKD~\cite{liu2019structured} & & & 67.45  \\
				+IFVD~\cite{wang2020intra} & & & 67.13   \\
				+CWD~\cite{shu2021channel} & & &  67.89  \\
				+DSD~\cite{feng2021double} & & &   67.34 \\
				+CIRKD~\cite{yang2022cross} & & &  68.04 \\
				+APD~\cite{tian2022adaptive}& & &  67.84 \\
				+Af-DCD~\cite{fan2023augmentation}& &  & \underline{68.14} \\
				+LAD~\cite{liu2024rethinking} & & &  67.48  \\
				+CIRKDV2 (Ours) & & &  \textbf{68.91}  \\
				\hline
				S: PSPNet-MobileViT-XXS &\multirow{10}{*}{1.8M} &\multirow{10}{*}{452.7G} & 65.48  \\
				+SKD~\cite{liu2019structured} & & &  66.84   \\
				+IFVD~\cite{wang2020intra} & & & 66.52   \\
				+CWD~\cite{shu2021channel} & & & 66.28   \\
				+DSD~\cite{feng2021double} & & & 66.74   \\
				+CIRKD~\cite{yang2022cross} & & & 67.45  \\
				+APD~\cite{tian2022adaptive}& & & 66.63 \\
				+Af-DCD~\cite{fan2023augmentation}& &  & \underline{67.54} \\
				+LAD~\cite{liu2024rethinking} & & & 66.56   \\
				+CIRKDV2 (Ours) & & & \textbf{68.45}  \\
				\hline
				
		\end{tabular}}
		
		\label{citys_vit} 
	\end{table}

	\begin{table}[tbp]
		\centering
		\caption{Performance comparison with the state-of-the-art distillation methods over SegFormer networks on Cityscapes.}
		\resizebox{1.\linewidth}{!}{
			\begin{tabular}{l|c|c|cc}  
				\hline
				Method&Params&FLOPs& mIoU (\%) \\ 
				\hline
				T: SegFormer-MiT-B4 &64.0M &1,229.8G & 80.38   \\
				\hline
				S: SegFormer-MiT-B0 &\multirow{10}{*}{3.7M} &\multirow{10}{*}{121.7G} & 74.12  \\
				+SKD~\cite{liu2019structured} & & &  74.35  \\
				+IFVD~\cite{wang2020intra} & & &  74.18  \\
				+CWD~\cite{shu2021channel} & & & 73.86   \\
				+DSD~\cite{feng2021double} & & & 74.70   \\
				+CIRKD~\cite{yang2022cross} & & & \underline{74.87}  \\
				+APD~\cite{tian2022adaptive}& & & 74.83 \\
				+Af-DCD~\cite{fan2023augmentation}& &  & 74.52  \\
				+LAD~\cite{liu2024rethinking} & & &  74.82  \\
				+CIRKDV2 (Ours) & & &  \textbf{75.52}  \\
				\hline
				
		\end{tabular}}
		
		\label{citys_segformer} 
	\end{table}

	\begin{figure}[tbp]  
		\centering 
		\includegraphics[width=1\linewidth]{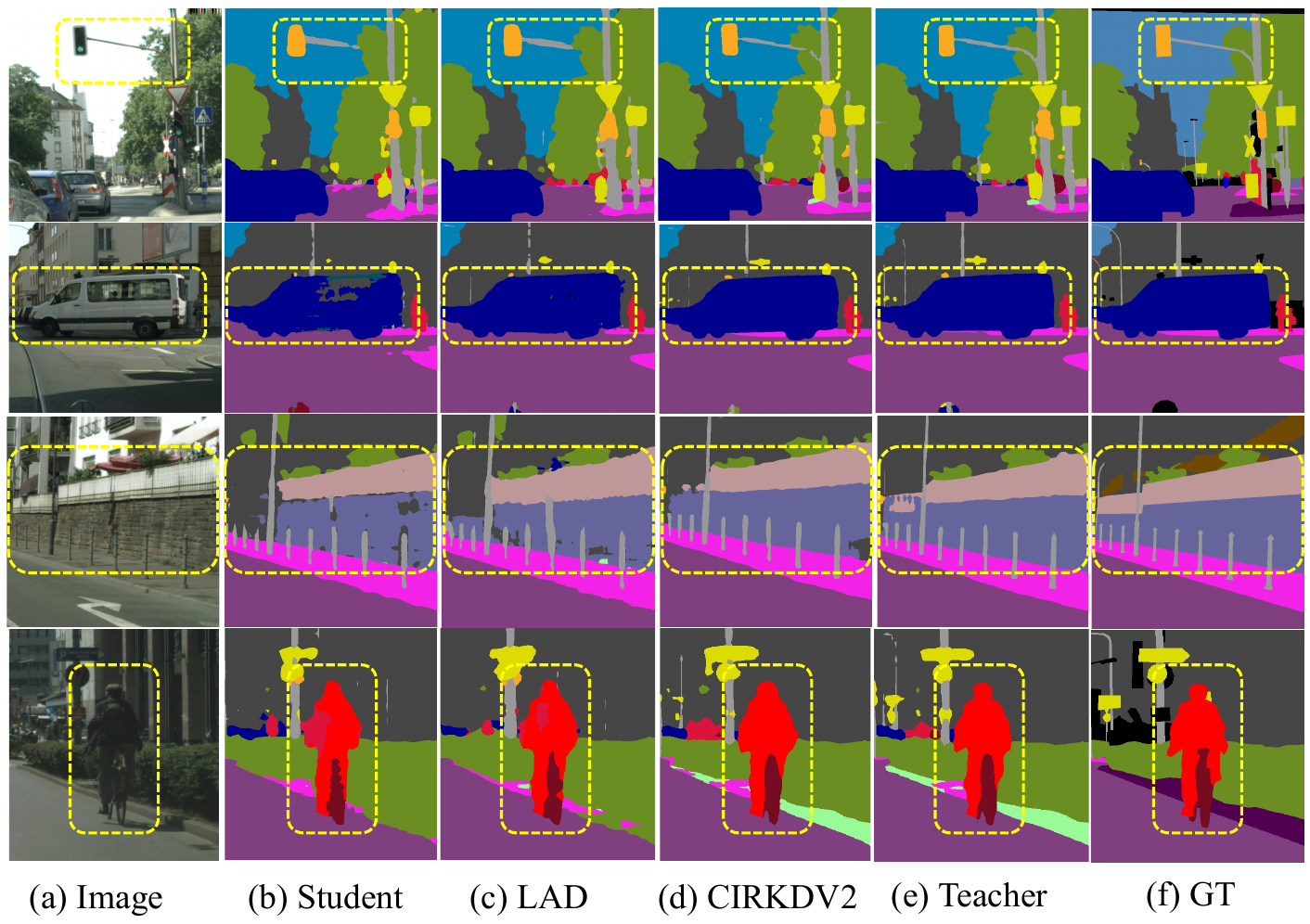}
		\caption{Qualitative segmentation results on the validation set of Cityscapes using
			the DeepLabV3-MobileNetV3-Large network: (a) raw images, (b) the
			original student network without KD,
			(c) LAD, (d) our CIRKDV2, (e) Teacher,  and (f) ground truth.} 
		\label{Qualitative_citys}
	\end{figure}
	
		\begin{figure}[!t]
			\centering
			\subfloat[LAD~\cite{liu2024rethinking}]{\includegraphics[width=0.2352\textwidth]{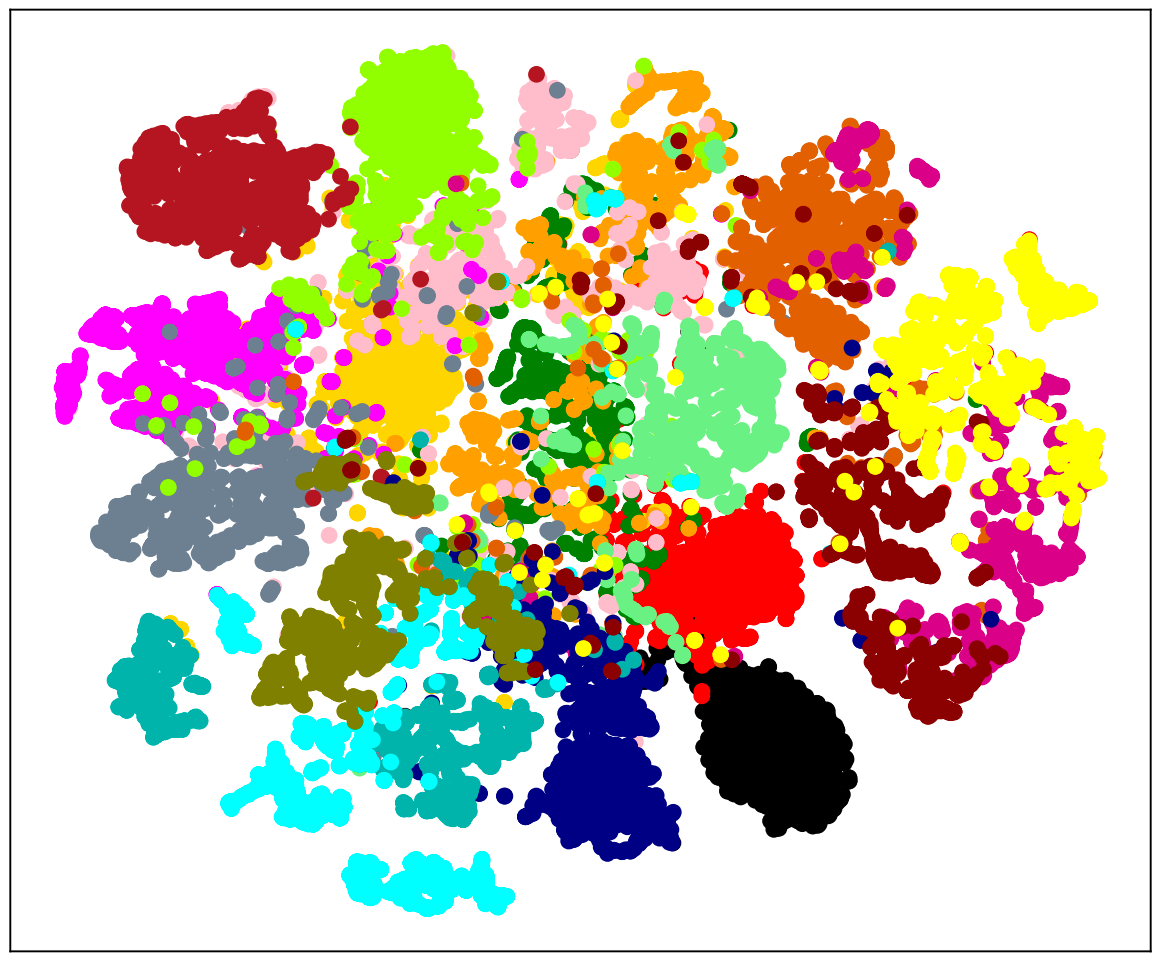}%
				\label{SMG}}
			\hfil
			\subfloat[Our CIRKDV2]{\includegraphics[width=0.2348\textwidth]{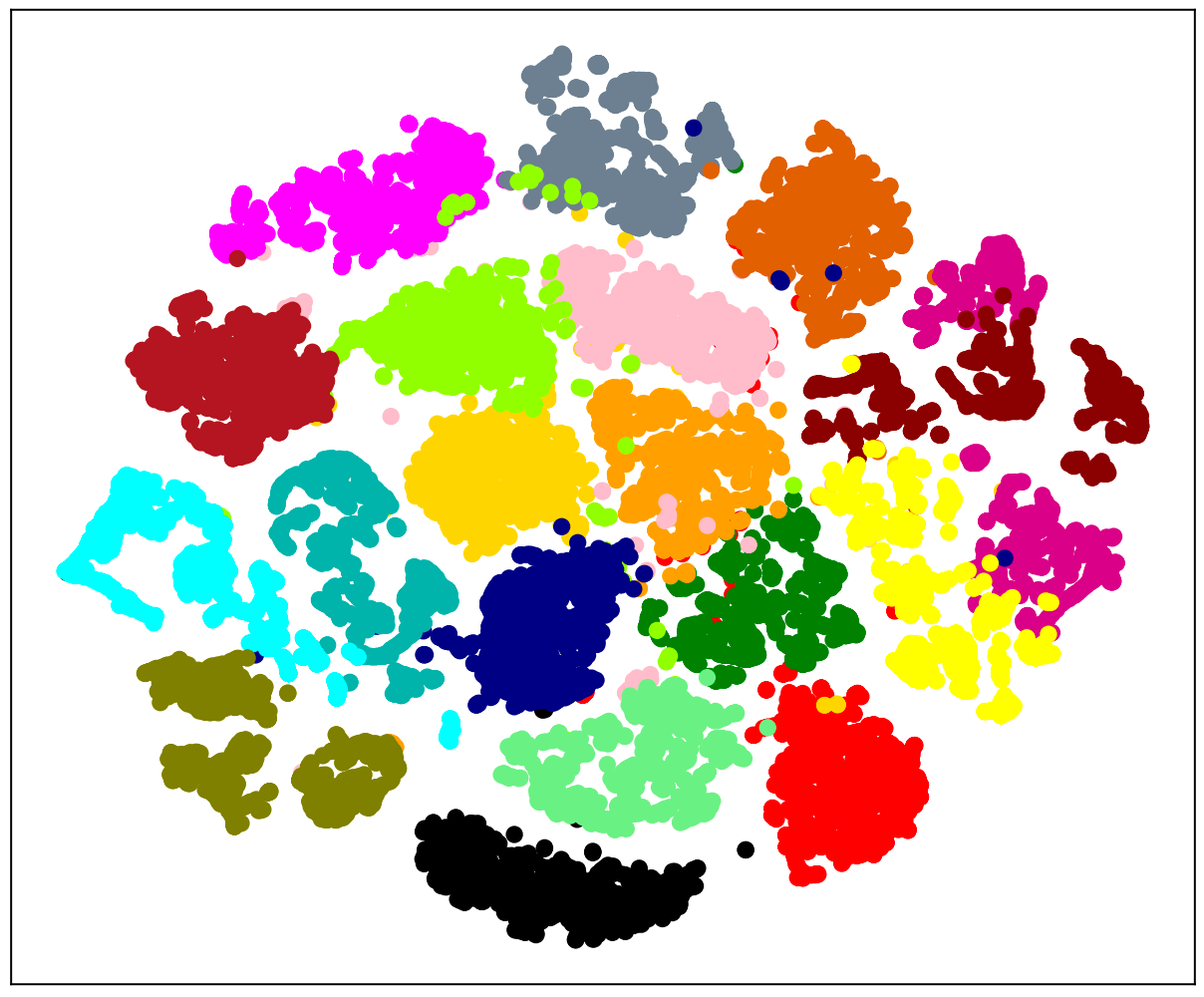}%
				\label{Update}}
			\caption{T-SNE visualization of learned feature embeddings on the validation set of Cityscapes over the DeepLabV3-ResNet18 network trained with LAD (\emph{left}) and our proposed CIRKDV2 (\emph{right}).}
			\label{tsne}
		\end{figure}
	
	\textbf{Hyper-parameters setup.} The hyper-parameters are mainly from the pixel, region, and channel queues. For the pixel queue, we set the queue size $N_{p}=20K$ for each class and enqueue $V_p=16$ pixels per class for each image.  For the region queue, we set the queue size $N_{r}=2K$ for each class. The channel queue size is $N_{c}=10K$, and we enqueue $V_c=256$ channels per iteration. For each mini-batch, we sample $K_{p}=4,096$ pixel embeddings from the pixel queue, $K_{r}=1,024$ region embeddings from the region queue, and $K_{c}=2,048$ channel embeddings from the channel queue to compute similarity matrices.
	
	\subsection{Results on Cityscapes}
	In Table~\ref{citys}, we compare our proposed CIRKDV2 against popular distillation methods on Cityscapes over CNN-based segmentation architectures. We can observe that all segmentation distillation methods improve student networks under the teacher's supervision. CIRKDV2 achieves the best segmentation performance across various student networks with homogeneous or heterogeneous architecture styles. Notice that LAD~\cite{liu2024rethinking}, a feature-based distillation method, is often superior in the homogeneous architecture pair like ResNet-101 and ResNet-18, but inferior in the heterogeneous architecture pair like ResNet-101 and MobileNetV3. By contrast, relation-based methods, like CIRKD~\cite{yang2022cross}, Af-DCD~\cite{fan2023augmentation}, and CIRKDV2, do not rely on architecture-specific cues. Moreover, CIRKDV2 outperforms the previous state-of-the-art CIRKD, Af-DCD, and LAD methods with 0.70\%, 0.88\%, and 0.98\% improvements on average across four CNN-based models, respectively. Table~\ref{citys_vit} shows the distillation results on MobileViT-XXS, a light-weight vision transformer for mobile devices. CIRKDV2 obtains 0.94\%, 0.84\%, and 1.66\% mIoU gains on average compared to CIRKD, Af-DCD, and LAD, respectively. As shown in Table~\ref{citys_segformer}, we further evaluate distillation methods on SegFormer. CIRKDV2 achieves the best performance and surpasses the second-best CIRKD by 0.65\%. The results demonstrate that CIRKDV2 can also bring remarkable improvements on transformer architectures. The superiority of CIRKDV2 may be attributed to distilling more comprehensive knowledge than previous feature- and relation-based methods.
	
	\begin{table}[tbp]
		\centering
		\caption{Performance comparison with the state-of-the-art distillation methods over various segmentation networks on ADE20K.}
		\resizebox{1.\linewidth}{!}{
			\begin{tabular}{l|c|c|cc}  
				\hline
				Method&Params&FLOPs& mIoU (\%) \\ 
				\hline
				T: DeepLabV3-ResNet-101 &79.2M &575.0G & 43.83  \\
				\hline
				S: DeepLabV3-ResNet-18 &\multirow{10}{*}{22.8M} &\multirow{10}{*}{167.5G} & 36.92  \\
				+SKD~\cite{liu2019structured} & & &  38.12  \\
				+IFVD~\cite{wang2020intra} & & & 37.67   \\
				+CWD~\cite{shu2021channel} & & &  38.96  \\
				+DSD~\cite{feng2021double} & & & 38.53   \\
				+CIRKD~\cite{yang2022cross} & & &  39.05  \\
				+APD~\cite{tian2022adaptive}& & & 38.55 \\
				+Af-DCD~\cite{fan2023augmentation}& &  & 39.21 \\
				+LAD~\cite{liu2024rethinking} & & &  \underline{39.37}  \\
				+CIRKDV2 (Ours) & & &  \textbf{39.82}  \\
				\hline
				S: UperNet-ResNet-18 &\multirow{10}{*}{16.6M} &\multirow{10}{*}{85.5G} &  34.37 \\
				+SKD~\cite{liu2019structured} & & &  35.25   \\
				+IFVD~\cite{wang2020intra} & & & 35.08   \\
				+CWD~\cite{shu2021channel} & & & 35.86   \\
				+DSD~\cite{feng2021double} & & &  35.33  \\
				+CIRKD~\cite{yang2022cross} & & & 35.81 \\
				+APD~\cite{tian2022adaptive}& & & 35.10  \\
				+Af-DCD~\cite{fan2023augmentation}& &  & 36.36 \\
				+LAD~\cite{liu2024rethinking} & & &  \underline{36.38}  \\
				+CIRKDV2 (Ours) & & &  \textbf{36.87}  \\
				\hline
				S: DeepLabV3-MobileNetV3-Large &\multirow{10}{*}{6.8M} &\multirow{10}{*}{49.0G} &  32.83 \\
				+SKD~\cite{liu2019structured} & & &  33.78  \\
				+IFVD~\cite{wang2020intra} & & & 34.21   \\
				+CWD~\cite{shu2021channel} & & &  34.18  \\
				+DSD~\cite{feng2021double} & & &  33.64  \\
				+CIRKD~\cite{yang2022cross} & & &  \underline{35.15} \\
				+APD~\cite{tian2022adaptive}& & & 34.53 \\
				+Af-DCD~\cite{fan2023augmentation}& &  & 35.12 \\
				+LAD~\cite{liu2024rethinking} & & &  34.61  \\
				+CIRKDV2 (Ours) & & &   \textbf{36.14} \\
				\hline
				S: PSPNet-MobileNetV3-Large &\multirow{10}{*}{6.1M} &\multirow{10}{*}{15.2G} & 33.63    \\
				+SKD~\cite{liu2019structured} & & &  34.82  \\
				+IFVD~\cite{wang2020intra} & & & 34.32   \\
				+CWD~\cite{shu2021channel} & & &  34.93  \\
				+DSD~\cite{feng2021double} & & &  33.90   \\
				+CIRKD~\cite{yang2022cross} & & &  35.27 \\
				+APD~\cite{tian2022adaptive}& & & 35.21 \\
				+Af-DCD~\cite{fan2023augmentation}& &  &  \underline{35.41}\\
				+LAD~\cite{liu2024rethinking} & & &  34.85 \\
				+CIRKDV2 (Ours) & & &   \textbf{36.01}  \\
				\hline
				
		\end{tabular}}
		
		\label{ade20k} 
	\end{table}

	\begin{table}[tbp]
		\centering
		\caption{Performance comparison with the state-of-the-art distillation methods over various segmentation networks on Pascal VOC.}
		\resizebox{1.\linewidth}{!}{
			\begin{tabular}{l|c|c|cc}  
				\hline
				Method&Params&FLOPs& mIoU (\%) \\ 
				\hline
				T: DeepLabV3-ResNet-101 &79.2M &575.0G & 77.80  \\
				\hline
				S: DeepLabV3-MobileNetV3-Small &\multirow{10}{*}{3.3M} &\multirow{10}{*}{28.1G} & 62.45   \\
				+SKD~\cite{liu2019structured} & & &  63.49  \\
				+IFVD~\cite{wang2020intra} & & &   63.33 \\
				+CWD~\cite{shu2021channel} & & &  63.67  \\
				+DSD~\cite{feng2021double} & & &  63.51  \\
				+CIRKD~\cite{yang2022cross} & & &  63.82  \\
				+APD~\cite{tian2022adaptive}& & & 63.61 \\
				+Af-DCD~\cite{fan2023augmentation}& &  & \underline{64.03} \\
				+LAD~\cite{liu2024rethinking} & & & 63.55  \\
				+CIRKDV2 (Ours) & & &  \textbf{64.67}  \\
				\hline
				S: DeepLabV3-MobileNetV3-Large &\multirow{10}{*}{6.8M} &\multirow{10}{*}{49.0G} & 69.33   \\
				+SKD~\cite{liu2019structured} & & & 70.57   \\
				+IFVD~\cite{wang2020intra} & & &   70.84 \\
				+CWD~\cite{shu2021channel} & & &  71.15  \\
				+DSD~\cite{feng2021double} & & &  70.66  \\
				+CIRKD~\cite{yang2022cross} & & & 71.24  \\
				+APD~\cite{tian2022adaptive}& & & 71.19  \\
				+Af-DCD~\cite{fan2023augmentation}& &  & \underline{71.32} \\
				+LAD~\cite{liu2024rethinking} & & &   71.08 \\
				+CIRKDV2 (Ours) & & &  \textbf{71.90}  \\
				\hline
				S: PSPNet-MobileNetV3-Small &\multirow{10}{*}{2.6M} &\multirow{10}{*}{7.4G} &  61.92 \\
				+SKD~\cite{liu2019structured} & & &  62.45  \\
				+IFVD~\cite{wang2020intra} & & &   62.51 \\
				+CWD~\cite{shu2021channel} & & &  63.12  \\
				+DSD~\cite{feng2021double} & & &  62.79  \\
				+CIRKD~\cite{yang2022cross} & & &  63.22 \\
				+APD~\cite{tian2022adaptive}& & & 62.43 \\
				+Af-DCD~\cite{fan2023augmentation}& &  & \underline{63.45} \\
				+LAD~\cite{liu2024rethinking} & & & 63.42   \\
				+CIRKDV2 (Ours) & & &   \textbf{63.84} \\
				\hline
				S: PSPNet-MobileNetV3-Large &\multirow{10}{*}{6.1M} &\multirow{10}{*}{15.2G} &  68.77   \\
				+SKD~\cite{liu2019structured} & & &  70.64  \\
				+IFVD~\cite{wang2020intra} & & &  70.45  \\
				+CWD~\cite{shu2021channel} & & &  70.28  \\
				+DSD~\cite{feng2021double} & & &  69.74   \\
				+CIRKD~\cite{yang2022cross} & & &  \underline{70.73} \\
				+APD~\cite{tian2022adaptive}& & & 70.34 \\
				+Af-DCD~\cite{fan2023augmentation}& &  & 70.62 \\
				+LAD~\cite{liu2024rethinking} & & &  70.06  \\
				+CIRKDV2 (Ours) & & &   \textbf{71.55}  \\
				\hline
				
		\end{tabular}}
		
		\label{pascal_voc} 
	\end{table}
	
	\begin{figure}[tbp]  
		\centering 
		\includegraphics[width=1\linewidth]{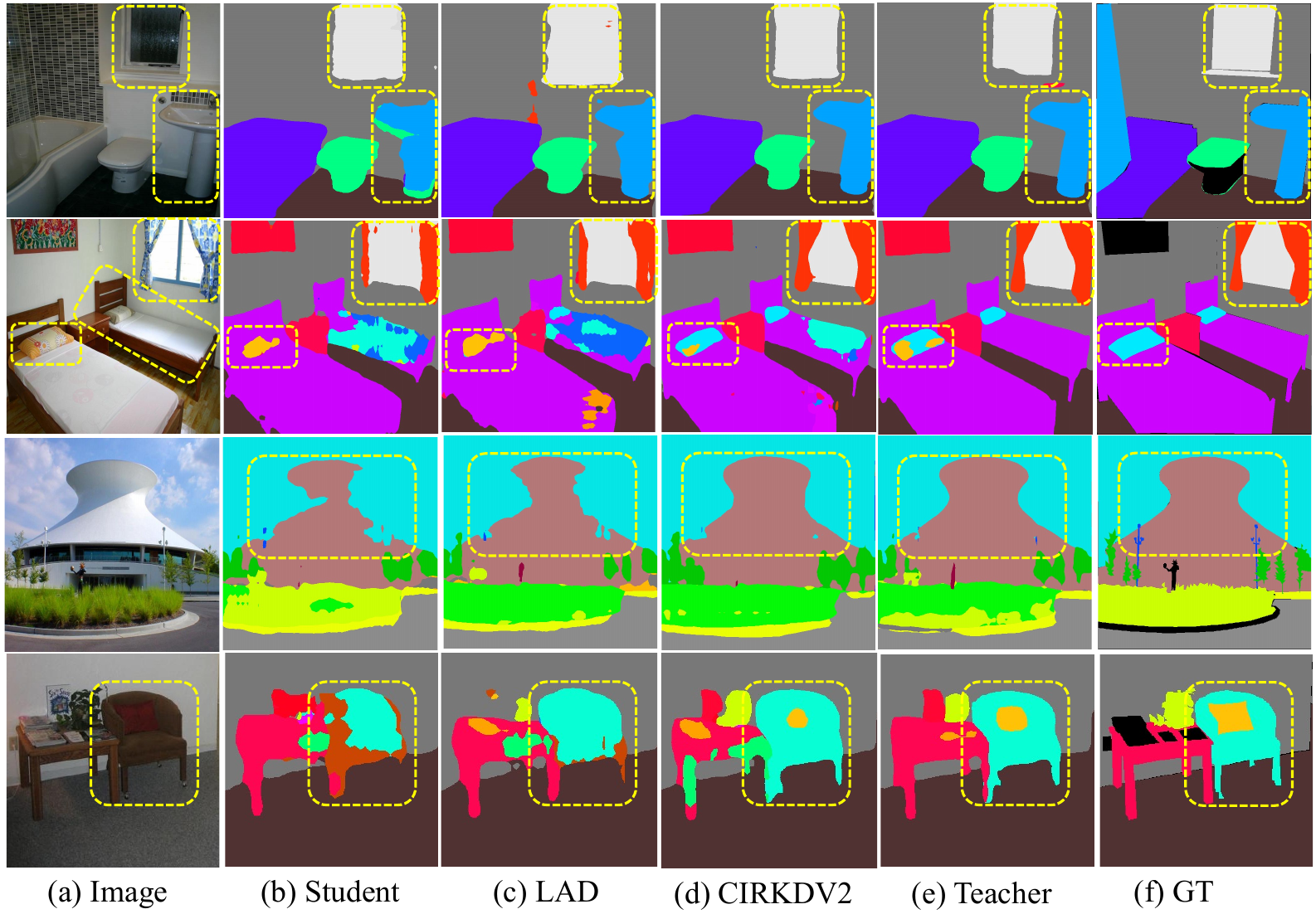}
		\caption{Qualitative segmentation results on the validation set of ADE20K using
			the DeepLabV3-MobileNetV3-Large network: (a) raw images, (b) the
			original student network without KD,
			(c) LAD, (d) our CIRKDV2, (e) Teacher, and (f) ground truth.} 
		\label{Qualitative}
	\end{figure}
	
	\begin{figure}[tbp]  
		\centering 
		\includegraphics[width=1\linewidth]{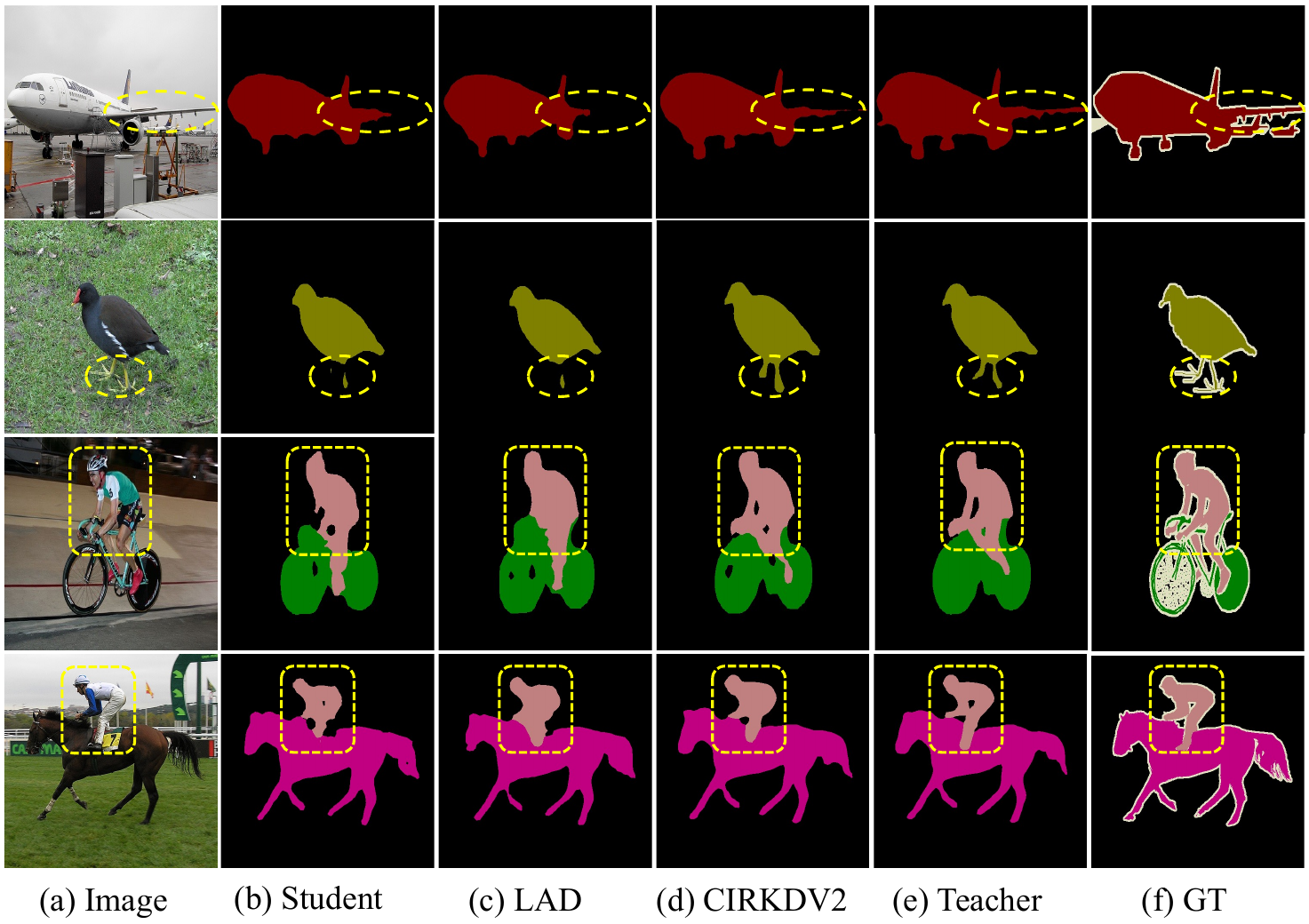}
		\caption{Qualitative segmentation results on the validation set of Pascal VOC using
			the DeepLabV3-MobileNetV3-Large network: (a) raw images, (b) the
			original student network without KD,
			(c) LAD, (d) our CIRKDV2, (e) Teacher, and (f) ground truth.} 
		\label{Qualitative_voc}
	\end{figure}

	\begin{table}[tbp]
		\centering
		\caption{Performance comparison with the state-of-the-art distillation methods over various segmentation networks on COCO-Stuff-164K.}
		\resizebox{1.\linewidth}{!}{
			\begin{tabular}{l|c|c|cc}  
				\hline
				Method&Params&FLOPs& mIoU (\%) \\ 
				\hline
				T: DeepLabV3-ResNet-101 &79.2M &575.0G&  38.48 \\
				\hline
				S: DeepLabV3-ResNet-18 &\multirow{10}{*}{22.8M} &\multirow{10}{*}{167.5G} & 32.65  \\
				+SKD~\cite{liu2019structured} & & & 33.27   \\
				+IFVD~\cite{wang2020intra} & & &   33.30 \\
				+CWD~\cite{shu2021channel} & & &  33.81  \\
				+DSD~\cite{feng2021double} & & &  33.26  \\
				+CIRKD~\cite{yang2022cross} & & &  33.92  \\
				+APD~\cite{tian2022adaptive}& & & 33.71 \\
				+Af-DCD~\cite{fan2023augmentation}& &  & 33.86 \\
				+LAD~\cite{liu2024rethinking} & & & \underline{33.95}   \\
				+CIRKDV2 (Ours) & & &  \textbf{34.42}  \\
				\hline
				S: PSPNet-MobileNetV3-Small &\multirow{10}{*}{2.6M} &\multirow{10}{*}{7.4G} & 26.48   \\
				+SKD~\cite{liu2019structured} & & &  27.52  \\
				+IFVD~\cite{wang2020intra} & & & 27.15   \\
				+CWD~\cite{shu2021channel} & & &  27.18  \\
				+DSD~\cite{feng2021double} & & &  27.42  \\
				+CIRKD~\cite{yang2022cross} & & &  27.66\\
				+APD~\cite{tian2022adaptive}& & & 27.28  \\
				+Af-DCD~\cite{fan2023augmentation}& &  & \underline{27.71} \\
				+LAD~\cite{liu2024rethinking} & & &  27.33  \\
				+CIRKDV2 (Ours) & & &  \textbf{28.28}  \\
				\hline
				S: DeepLabV3-MobileNetV3-Small &\multirow{10}{*}{3.3M} &\multirow{10}{*}{28.1G} & 26.04   \\
				+SKD~\cite{liu2019structured} & & &  26.72  \\
				+IFVD~\cite{wang2020intra} & & & 26.46   \\
				+CWD~\cite{shu2021channel} & & &  27.12  \\
				+DSD~\cite{feng2021double} & & &  26.73  \\
				+CIRKD~\cite{yang2022cross} & & &  26.83 \\
				+APD~\cite{tian2022adaptive}& & & 26.70 \\
				+Af-DCD~\cite{fan2023augmentation}& &  &  \underline{27.25}\\
				+LAD~\cite{liu2024rethinking} & & & 27.11   \\
				+CIRKDV2 (Ours) & & &   \textbf{27.66} \\
				\hline
				S: DeepLabV3-MobileNetV3-Large &\multirow{10}{*}{6.8M} &\multirow{10}{*}{49.0G} &  30.31   \\
				+SKD~\cite{liu2019structured} & & &  30.83  \\
				+IFVD~\cite{wang2020intra} & & & 30.99  \\
				+CWD~\cite{shu2021channel} & & &   31.24 \\
				+DSD~\cite{feng2021double} & & &   30.81  \\
				+CIRKD~\cite{yang2022cross} & & &  \underline{31.62} \\
				+APD~\cite{tian2022adaptive}& & & 31.45  \\
				+Af-DCD~\cite{fan2023augmentation}& &  & 31.59 \\
				+LAD~\cite{liu2024rethinking} & & & 31.08  \\
				+CIRKDV2 (Ours) & & &   \textbf{32.14}  \\
				\hline
				
		\end{tabular}}
		
		\label{coco_stuff_164k} 
	\end{table}
	
	\begin{figure}[tbp]  
		\centering 
		\includegraphics[width=1\linewidth]{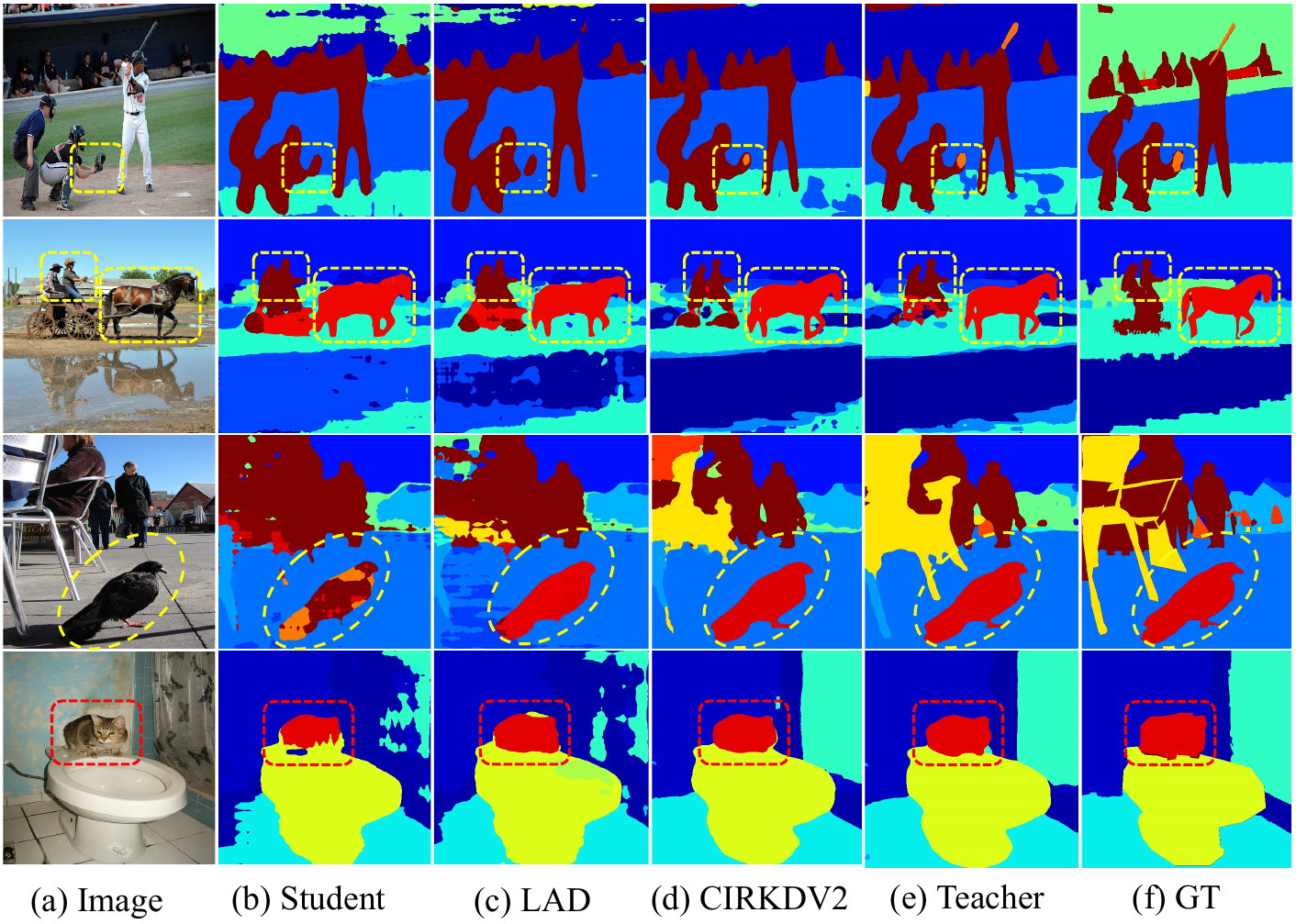}
		\caption{Qualitative segmentation results on the validation set of COCO-Stuff-164K using
			the DeepLabV3-MobileNetV3-Large network: (a) raw images, (b) the
			original student network without KD,
			(c) LAD, (d) our CIRKDV2, (e) Teacher, and (f) ground truth.} 
		\label{Qualitative_coco}
	\end{figure}

	As illustrated in Figure~\ref{cityscape_iou}, we also show the performance of individual class IoU scores over the student network. We can observe that our CIRKDV2 achieves better class IoU scores than baseline (W/O distillation) and LAD consistently, especially for those categories with low IoU scores. For example, CIRKDV2 obtains 2.08\%, 5.18\%, and 5.81\%  gains on \textit{wall}, \textit{fence}, and \textit{truck} than LAD, respectively. We further show the qualitative segmentation results visually in Figure~\ref{Qualitative_citys}. We can observe that our CIRKDV2 produces more consistent semantic labels with the ground truth than baseline and LAD, indicating more meaningful pixel semantics are captured.
	
	T-SNE visualization of learned feature embeddings on the student network by LAD and our proposed CIRKDV2 is shown in Figure~\ref{tsne}. Compared to LAD, the network trained by CIRKDV2 shows a more discriminative pixel-wise semantic feature space. The visual result suggests that learning cross-image relations with pixel-to-pixel, pixel-to-region, and channel-to-channel dimensions from the teacher network can guide the student to learn comprehensive knowledge, helping the student achieve better intra-class compactness and inter-class separability, thus improving segmentation performance.

	\subsection{Results on ADE20K}
	In Table~\ref{ade20k}, we evaluate various distillation methods on ADE20K. ADE20K is a popular semantic segmentation benchmark and more challenging than Cityscapes. CIRKDV2 also achieves the best performance across various segmentation frameworks. It surpasses previous state-of-the-art CIRKD, Af-DCD, and LAD methods by 0.89\%, 0.69\%, and 0.91\% on average across four networks, respectively. The results manifest that CIRKDV2 can also work well on more diverse scenes.
	
	\subsection{Results on Pascal VOC}
	Beyond scene-parsing datasets, we also evaluate our CIRKDV2 on Pascal VOC, a representative visual object segmentation dataset. As shown in Table~\ref{pascal_voc}, CIRKDV2 achieves the best performance compared to other segmentation KD approaches. 
	Compared to the previous state-of-the-art CIRKD, Af-DCD, and LAD methods, CIRKDV2 results in 0.74\%, 0.64\%, and 0.96\% improvements across four segmentation models on average, respectively. The results indicate that CIRKDV2 can lead to accurate semantic masks on visual object segmentation.
	
	\subsection{Results on COCO-Stuff-164K}
	As shown in Table~\ref{coco_stuff_164k}, we compare various distillation methods on COCO-Stuff-164K. COCO-Stuff-164K is a large-scale semantic segmentation dataset and has more complex scenes and objects. CIRKDV2 performs the best consistently among the existing segmentation distillation methods. It brings 0.62\%, 0.52\%, and 0.76\% mIoU gains over CIRKD, Af-DCD, and LAD methods on average across four networks, respectively. The promising results demonstrate the good scalability of our CIRKDV2 to work reasonably well on large-scale and complex scenes.
	
	We follow the MobileSAM~\cite{zhang2023faster} setup to apply CIRKDV2 to distill a SAM model with a lightweight image encoder ViT-Tiny. The teacher is a pretrained SAM model with a high-performance ViT-Large image encoder. As shown in Table \ref{sam_distill}, CIRKDV2 achieves 3.16\% and 1.89\% mIoU improvements compared to the baseline and MobileSAM on the challenging COCO-Stuff-164K, respectively. The results demonstrate that CIRKDV2 can also be effectively extended to SAM foundation models.
	
	\begin{table}[tbp]
		\centering
		\caption{Distillation experiments on SAM model on COCO-Stuff-164K.}
		\begin{tabular}{l|c}
			\hline
			Method& mIoU (\%) \\ 
			\hline
			T: SAM-ViT-Large  &  42.82 \\
			\hline
			S: SAM-ViT-Tiny  & 27.64  \\
			+MobileSAM~\cite{zhang2023faster}& 28.91 \\
			+CIRKDV2 & \textbf{30.80} \\
			\hline
		\end{tabular}
		\label{sam_distill} 
	\end{table}
	
	\begin{table}[tbp]
		\centering
		\caption{Comparison of training efficiency over DeepLabV3-ResNet18 on ADE20K.}
		
		\begin{tabular}{l|cccc}  
			\hline
			Method& Time & GPU Memory &mIoU (\%) \\ 
			\hline
			+Af-DCD~\cite{fan2023augmentation}& 16s & 9.8GB  & 39.21 \\
			+LAD~\cite{liu2024rethinking} & 12s & 8.5GB&  39.37  \\
			+CIRKDV2 (Ours) & 17s&  11.7GB  &  \textbf{39.82}  \\
			\hline
		\end{tabular}
		
		\label{training_complexity} 
	\end{table}

\begin{table}[htbp]
	\centering
	\caption{Ablation study of Contextual FD and C2C distillation on SegFormer.}
	\resizebox{1.\linewidth}{!}{
		\begin{tabular}{l|ccc}
			\hline
			Student: SegFormer & B0 &  B0& B1  \\
			Dataset & Citys & ADE20K &  ADE20K\\
			\hline
			W/O Distillation & 74.12 & 33.83& 38.12\\
			+Contextual FD(W/O Nonlinear Branch) &74.67&35.18 & 38.73 \\
			+Contextual FD & 74.75 &35.39 & 38.94\\
			+Channel FD  & 74.48 & 34.56& 38.40\\
			+mini-batch C2C & 74.76 &35.89 & 39.34\\
			+memory-based C2C & 74.92 &36.28 & 39.62\\
			All & \textbf{75.24}&\textbf{36.74} &\textbf{40.18} \\
			\hline
	\end{tabular}}
	\label{ablation_segformer} 
\end{table}

	\subsection{Comparison of training complexity}
	As shown in Table~\ref{training_complexity}, we compare our CIRKDV2 with Af-DCD and LAD methods, which represent the state-of-the-art relation and feature distillation works, respectively. The training time is recorded from one training step with a batch size of 16 on 8 NVIDIA RTX 4090 GPUs. LAD achieves the best training efficiency because it aligns feature maps without sophisticated relational modeling between teacher and student. Both Af-DCD and CIRKDV2 are relation-based distillation methods. Compared to Af-DCD, CIRKDV2 needs extra 6.25\% training time and 19.39\% GPU memory costs but results in the best 39.82\% mIoU. The good performance justifies the minor increase of training costs. Notice that all distillation methods do not change the student network architecture, therefore, the inference time and memory are the same as the original baseline network without distillation.
	
	\section{Ablation Study and Parameter Analysis}
	We conduct thorough ablation experiments of our proposed CIRKDV2 on ADE20K, a standard yet challenging benchmark for semantic segmentation. For all experiments in this section, we choose DeepLabV3-ResNet-101 as the teacher and DeepLabV3-ResNet-18 as the student by default.
	
	\begin{table*}[t]
		\centering
		\caption{Ablation study of distillation loss terms on ADE20K.}
		
		\begin{tabular}{l|c|c|ccccc|ccccc|c|c}  
			\hline 
			Loss&$L_{task}$& \multicolumn{13}{c}{Various distillation loss combinations of CIRKDV2}\\ 
			\hline
			$L_{kd}$&-&\checkmark&\checkmark&\checkmark&\checkmark&\checkmark&\checkmark&\checkmark&\checkmark&\checkmark&\checkmark&\checkmark&\checkmark&\checkmark \\
			\hline
			$L_{batch\_p2p}$&-&-&\checkmark&-&-&-&\checkmark&-&-&-&-&-&-&\checkmark\\
			$L_{batch\_c2c}$&-&-&-&\checkmark&-&\checkmark&\checkmark&-&-&-&-&-&-&\checkmark\\
			\hline
			$L_{memory\_p2p}$&-&-&-&- &-&-&-&\checkmark&-&-&-&\checkmark&-&\checkmark\\
			$L_{memory\_p2r}$&-&-&-&-&-&-&-&-&\checkmark&-&-&\checkmark&-&\checkmark\\
			$L_{memory\_c2c}$&-&-&-&-&-&-&-&-&-&\checkmark&\checkmark&\checkmark&-&\checkmark\\
			\hline
			$L_{channel\_fd}$&-&-&-&-&\checkmark&\checkmark&\checkmark&-&-&-&\checkmark&\checkmark&-&\checkmark\\
			$L_{context\_fd}$&-&-&-&-&-&-&-&-&-&-&-&-&\checkmark&\checkmark\\
			\hline
			mIoU (\%) & 36.92 &37.38&38.24&38.01&37.68&38.56&38.82&38.77&38.62&38.56&39.06&39.32&38.53&\textbf{39.82}\\
			\hline
			Method & Baseline & KD &\multicolumn{5}{c|}{Mini-batch-based relational distillation} & \multicolumn{5}{c|}{Memory-based relational distillation} & Context & Overall \\
			\hline
		\end{tabular}
		\label{ablation}
		\vspace{-0.3cm}
	\end{table*}
	
	\subsection{Ablation study of loss terms} As shown in Table~\ref{ablation}, we conduct a comprehensive ablation study to examine the contribution of each distillation loss and various loss combinations. 
	
	\textbf{KD loss}. The conventional KD loss $L_{kd}$ improves the baseline $L_{task}$ loss by 0.46\%. This means that pixel-wise class probability distillation leads to a moderate improvement for semantic segmentation.  
	
	\textbf{Mini-batch-based relational distillation loss}. Applying mini-batch-based relational KD losses of $L_{batch\_p2p}$, $L_{batch\_c2c}$, and $L_{batch\_c2c}+L_{channel\_fd}$ leads to 1.32\%, 1.09\% and 1.64\% mIoU gains over the baseline without distillation, respectively. $L_{channel\_fd}$ boosts $L_{batch\_c2c}$ by 0.55\%, because it regularizes the semantics of student channels to be consistent with the teacher channels. The overall mini-batch-based relational distillation losses achieve a 1.9\% mIoU improvement over baseline. 
	
	\textbf{Memory-based relational distillation loss}. The memory-based relational distillation losses of $L_{memory\_p2p}$, $L_{memory\_p2r}$, $L_{memory\_c2c}$, $L_{memory\_c2c}+L_{channel\_fd}$ result in 1.39\%, 1.24\%, 1.18\%, and  1.68\% mIoU improvements, respectively. Similar to mini-batch-based channel-to-channel distillation, $L_{channel\_fd}$ can also facilitate the effectiveness of memory-based channel-to-channel distillation $L_{memory\_c2c}$ with a 0.50\% mIoU gain. Moreover, we found that channel-to-channel distillation is better than pixel-to-pixel and pixel-to-region distillation. This reveals that modeling channel interaction is more informative than pixel-to-pixel and pixel-to-region interactions. The overall memory-based relational distillation losses outperform baseline and mini-batch-based relational distillation by 2.40\% and 0.50\%, respectively. Compared to mini-batch-based distillation, memory-based distillation can capture broader relational dependencies from many more images, therefore bringing better segmentation performance.
	
	\textbf{Contextual feature distillation}. The loss $L_{context\_fd}$ achieves a 1.15\% mIoU increase over the traditional KD $L_{kd}$ while keeping the same inference cost as the original student. Moreover, combining cross-image relational distillation with contextual feature distillation maximizes the segmentation mIoU to 39.82\%, reducing the performance gap between teacher and student from 6.91\% to 4.01\%.
	
		\textbf{Channel feature distillation}. As shown in Table \ref{channel}, the ablation experiments on heterogeneous teacher-student pairs, such as ResNet$\rightarrow$MobileNetV3 and ResNet$\rightarrow$MobileViT, show consistent improvements of 1.18\% and 1.24\%, suggesting that the proposed design does not cause representation collapse or negative transfer. We further demonstrate that channel distillation leads to one-to-one semantic alignment between teacher and student channels. As shown in Appendix Figure R.2, we found that the student trained by $L_{channel\_fd}$ exhibits similar semantic correspondence with the teacher. By contrast, the independently trained student without $L_{channel\_fd}$ loss demonstrates different semantic patterns with the teacher. The qualitative results support the claimed channel semantic consistency.

		In Table \ref{ablation_segformer}, we systematically evaluate the individual contributions of Contextual FD, Channel FD, mini-batch C2C, and memory-based C2C on SegFormer architectures across Cityscapes and ADE20K. Specifically, on SegFormer-MiT-B0 with ADE20K, Contextual FD, Channel FD, mini-batch C2C, and memory-based C2C improve the baseline mIoU from 33.83\% to 35.39\%, 34.56\%, 35.89\%, and 36.28\%, respectively. Similar improvements are observed on Cityscapes and with SegFormer-MiT-B1 on ADE20K. The substantial gains retained after removing the branch support the effectiveness of the contextual feature alignment mechanism.
		
		\begin{table*}[t]
			\begin{minipage}{0.31\linewidth}
			\centering
				\caption{Ablation study of $L_{channel\_fd}$ over heterogeneous teacher-student pairs on Cityscapes.}
				\resizebox{1\textwidth}{!}{
				\begin{tabular}{l|c}
					\hline
					Method& mIoU (\%) \\ 
					\hline
					T: DeepLabV3-ResNet-101  &78.30 \\
					S: PSPNet-MobileNetV3-Small & 62.78  \\
					+$L_{channel\_fd}$ & \textbf{63.96}   \\
					\hline
					T: DeepLabV3-ResNet-101  & 78.30   \\
					S: PSPNet-MobileViT-XXS & 65.48 \\
					+$L_{channel\_fd}$ &\textbf{66.72} \\
					\hline
			\end{tabular}}
			\label{channel} 
			\end{minipage}
			\hfill
			\begin{minipage}{0.31\linewidth}  
			\centering
			\caption{Ablation study of context aggregation for feature distillation on ADE20K.}
			\resizebox{1\textwidth}{!}{
			\begin{tabular}{l|c}  
				\hline
				Method& mIoU (\%) \\ 
				\hline
				Baseline & 36.92 \\
				W/O contextual aggregation & 37.33  \\
				Equal contextual aggregation & 37.65\\
				Object contextual aggregation~\cite{yuan2020object} & 37.82 \\
				Attention contextual aggregation (Ours) & \textbf{38.53}  \\
				\hline
			\end{tabular}}
			
			\label{context_ablation}
			\vspace{-0.3cm}
				
			\end{minipage}
			\hfill
			\begin{minipage}{0.31\linewidth}  
			
			\centering
			\caption{Analysis of various sampling methods on ADE20K.}
			
			\resizebox{1\textwidth}{!}{
			\begin{tabular}{l|c|c}
				\hline
				Method& Complexity &mIoU (\%) \\ 
				\hline
				W/O sampling & $O(1)$ & 39.24 \\
				Uniform sampling & $O(1)$& 39.55 \\
				Random sampling & $O(1)$&  \textbf{39.82} \\
				Hardest sampling & $C(N_{p}+N_{r})+N_c$ & 39.44\\
				Semi-hard sampling & $C(N_{p}+N_{r})+N_c$ & 39.65 \\
				\hline
			\end{tabular}}
			\label{sampling} 
		\end{minipage}
		\end{table*}

	\subsection{Ablation of context aggregation for feature distillation}
	As shown in Table~\ref{context_ablation}, we conduct an ablation study of various context aggregation methods for feature distillation. The feature distillation without contextual aggregation only achieves a 0.41\% mIoU gain over baseline. This is because the traditional feature distillation ignores that the teacher pixel has a wider receptive field than the student pixel. Although contextual aggregation with equal weights improves the traditional feature distillation by 0.32\%, it neglects the correlation among contextual pixels.
	The object contextual aggregation~\cite{yuan2020object} absorbs object region representations to augment student pixels and boosts the equal contextual aggregation by 0.17\%. We think the reason why object region representations lead to a marginal improvement is that the object-level coarse-grained representations do not contribute to the optimization of pixel-wise feature distillation. Our method introduces the attention mechanism for weighted pixel-level contextual aggregation and achieves the best 1.61\% improvement over baseline. The results demonstrate that our method makes the student pixels learn better semantic information from the teacher. This is because our method widens the student receptive field and leads the student to learn matched pixel semantic information from the teacher.

	\subsection{Impact of the queue size}
	We investigate the impact of memory sizes of the pixel queue, region queue, and channel queue. As shown in Appendix Figure R.1, distillation performance increases as the sizes of the pixel queue, region queue, and channel queue grow. This is because a larger queue could provide more abundant and diverse embeddings for capturing long-range dependencies. The results also show that the distillation performance saturates at a certain memory capacity. In summary, the pixel, region, and channel queues occupy 368MB, 40MB, and 92MB memory costs, which only account for $<$1\% of the whole memory.
	
	In our design, the region queue stores features generated \textbf{by the frozen teacher network}. Since the teacher is fixed, the stored region embeddings remain semantically invariant throughout training, therefore the “staleness” problem does not exist. In addition, we use a class-wise queue and class-balanced sampling. This prevents frequent categories from dominating the relation construction and ensures that rare classes can still be sampled once their region embeddings are stored. As shown in Figure \ref{cityscape_iou} of our manuscript, the Cityscapes class-wise IoU analysis also shows gains than SOTA distillation method LAD on rare classes like wall, fence, and truck by  2.08\%, 5.18\% and  5.81\% mIoUs, respectively. This indicates that the class-wise region queue does not harm rare classes. Instead, it helps preserve teacher-provided prototypes and makes them available to provide stable information for later iterations.
	 
	We investigate whether a large memory queue is more effective than a smaller but more frequently refreshed queue. In Table \ref{queue_size_deeplab} and \ref{queue_size_segformer}, the experiments are conducted on both CNN-based (DeepLabV3-ResNet-18) and Transformer-based (SegFormer-MiT-B0) architectures across Cityscapes and ADE20K. For DeepLabV3-ResNet-18, the large queue (20K, 16) achieves 75.26\% and 38.77\% mIoU on Cityscapes and ADE20K, respectively, outperforming the smaller queue (5K, 40) by 0.22\% and 0.42\% mIoU. Similarly, for SegFormer-MiT-B0, the large queue achieves 75.35\% and 36.52\% mIoU, exceeding the smaller queue by 0.17\% and 0.25\% mIoU. The intermediate configurations also exhibit a consistent trend of improved performance with larger queue capacities. These results suggest that preserving diverse historical embeddings is more beneficial than increasing the queue refresh rate in our cross-image relational distillation framework.
	
	\begin{table}[htbp]
		\centering
		\caption{Analysis of pixel queue size with update frequency on DeepLabV3-ResNet-18.}
		\begin{tabular}{l|cccc}
			\hline
			Queue size  $N_p$ & 20K & 15K & 10K & 5K \\
			Enqueue size $V_p$ & 16 & 24 & 32 & 40  \\
			\hline
			Cityscapes mIoU (\%) & \textbf{75.26} & 75.18&75.12 & 75.04\\
			ADE20K mIoU (\%) &\textbf{38.77} & 38.68& 38.57& 38.35 \\
			\hline
		\end{tabular}
		\label{queue_size_deeplab} 
	\end{table}
	
	\begin{table}[htbp]
		\centering
		\caption{Analysis of pixel queue size with update frequency on SegFormer-MiT-B0.}
		\begin{tabular}{l|cccc}
			\hline
			Queue size  $N_p$ & 20K & 15K & 10K & 5K \\
			Enqueue size $V_p$ & 16 & 24 & 32 & 40  \\
			\hline
			Cityscapes mIoU (\%) & \textbf{75.35} & 75.34 & 75.26 & 75.18\\
			ADE20K mIoU (\%) &\textbf{36.52} & 36.50 & 36.42 & 36.27 \\
			\hline
		\end{tabular}
		\label{queue_size_segformer} 
	\end{table}

	\subsection{Impact of sampling methods}
	As shown in Table~\ref{sampling}, we examine the impact of various sampling methods for memory-based distillation. “W/O sampling” means that the queue size is equal to the number of contrastive samples. “Hardest sampling” represents that we sample the hardest contrastive instances from the queues. Some previous studies~\cite{schroff2015facenet,cai2020all} found that the hardest negatives may lead to bad local minima. To alleviate this problem, we attempt a semi-hard sampling strategy. We first collect the top 20\% hardest negatives from the queues and then randomly sample contrastive instances from them. We found that “W/O sampling” is harmful to the performance gain, since this way decreases the queue size and range for modeling dependencies. Other advanced sampling methods, like hardest and semi-hard  sampling, also do not produce improvements but introduce extra computation complexity. Finally, we choose random sampling, a simple yet effective way.
	
	In dense semantic segmentation, the memory bank is used to approximate the teacher's dataset-level relational distribution, not only to mine the hardest negatives. A large queue provides diverse cross-image semantic contexts, while random class-balanced sampling gives a stable estimate of this global distribution. \textbf{By contrast, hardest mining can overemphasize outliers, boundary pixels, mislabeled regions, or visually similar classes that are not reliable negatives.} This may explain why hardest and semi-hard sampling introduce extra computation but do not improve performance.

	\begin{figure}[!t]
		\centering
		\subfloat[Temperature $\tau$]{\includegraphics[width=0.239\textwidth]{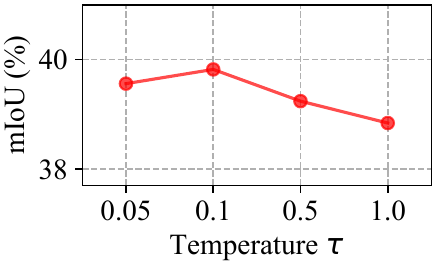}%
			\label{temperature}}
		\hfil
		\subfloat[Pixel-based $K_{p}$]{\includegraphics[width=0.234\textwidth]{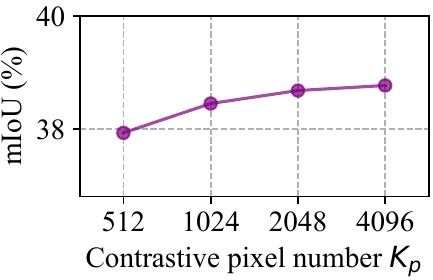}%
			\label{Pixel}}
		\hfil
		\subfloat[Region-based $K_{r}$]{\includegraphics[width=0.239\textwidth]{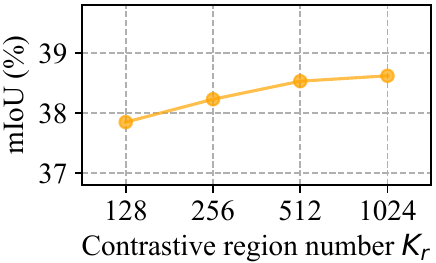}
			\label{Region}}
		\subfloat[Channel-based $K_{c}$]{\includegraphics[width=0.236\textwidth]{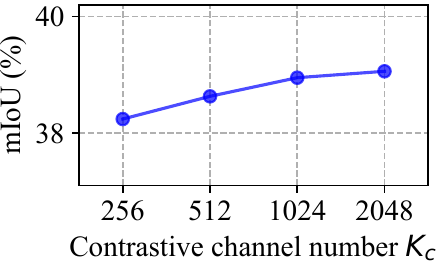}
			\label{Channel}}
		\caption{Impact of (a) the temperature $\tau$, (b) the number of contrastive pixel embeddings $K_{p}$, (c) the number of contrastive region embeddings $K_{r}$, and (d) the number of contrastive channel embeddings $K_{c}$ on ADE20K.}
		\label{params}
	\end{figure}
	
	\subsection{Impact of the temperature $\tau$}
	Temperature $\tau$ is used to calibrate the similarity distribution for relational distillation. A more significant temperature $\tau$ brings a smoother distribution. As shown in Figure~\ref{temperature}, we investigate the impact of $\tau$ and find $\tau=0.1$ is the best choice. A smaller $\tau$ than 0.1 may lead to under-smoothing, while a larger $\tau$ than 0.1 may result in over-smoothing, causing information loss of similarity distributions.

	\subsection{Impact of the number of contrastive embeddings}
	As shown in Figure~\ref{Pixel},~\ref{Region},~\ref{Channel}, we examine the number of contrastive embeddings to calculate pixel-to-pixel, pixel-to-region, and channel-to-channel similarity matrices. The distillation performance increases as $K_{p}$, $K_{r}$, and $K_{c}$ grow, because the similarity distribution with a larger dimension would encode broader pixel dependencies. The upper bound of distillation performance may saturate at $K_{p}=4,096$ for pixel-to-pixel distillation, $K_{r}=1,024$ for pixel-to-region distillation, and $K_{c}=2,048$ for channel-to-channel distillation.

	\begin{figure}[!t]
		\centering
		\subfloat[Mini-batch-based $\alpha$]{\includegraphics[width=0.234\textwidth]{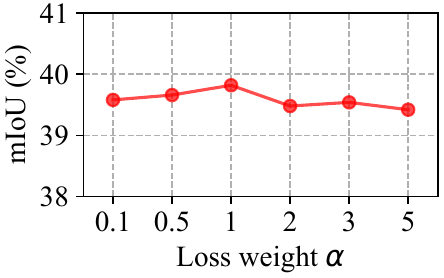}
			\label{alpha}}
		\hfil
		\subfloat[Memory-based $\beta$]{\includegraphics[width=0.234\textwidth]{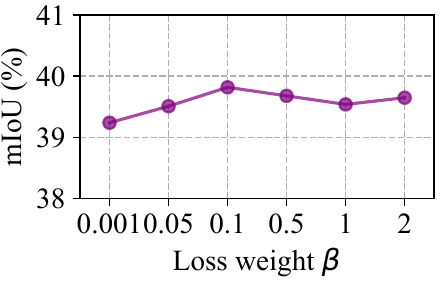}
			\label{beta}}
		\hfil
		\subfloat[Channel-based $\gamma$]{\includegraphics[width=0.238\textwidth]{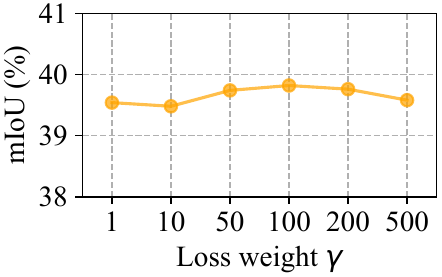}
			\label{gamma}}
		\subfloat[Context-based $\lambda$]{\includegraphics[width=0.238\textwidth]{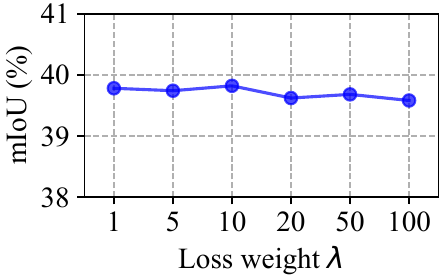}
			\label{lambda}}
		\caption{Sensitivity analyses of (a) mini-batch-based distillation loss weight $\alpha$, (b) memory-based distillation loss weight $\beta$, (c) channel distillation loss weight $\gamma$, and (d) contextual distillation loss weight $\lambda$ on ADE20K.}
		\label{loss_weights}
	\end{figure}
	
	\subsection{Sensitivity analyses of loss weights}
	As shown in Figure~\ref{loss_weights}, we analyze the sensitivity of the loss weights $\alpha$, $\beta$, $\gamma$, and $\lambda$, where we set $\alpha\in [0.1,5]$, $\beta\in [0.001,2]$, $\gamma\in [1,500]$, and $\lambda\in [1,100]$. For relational distillation weights $\alpha$ and $\beta$, we find that performance increases from $\alpha=0.1$ to $\alpha=1$, and from $\beta=0.001$ to $\beta=0.1$, reaching the best results when $\alpha=1$ and $\beta=0.1$. When $\alpha>1$ and $\beta>0.1$, the performance becomes suboptimal. We think that smaller $\alpha$ and $\beta$ reduce the contributions of relational distillation, while larger counterparts may lead to over-optimization and negative regularization. Moreover, we observe that $\gamma\in [50,200]$ and $\lambda\in [1,10]$ work well for feature distillation. Overall, loss weights are not sensitive within our recommended ranges.

	\section{Conclusion}
	This paper presents CIRKDV2, an improved cross-image relational KD method for efficient semantic segmentation. We construct pixel-to-pixel, pixel-to-region, and channel-to-channel relation modeling to transfer multi-dimensional structured knowledge for learning long-range pixel and channel dependencies. Moreover, we propose contextual feature distillation that first performs spatial context aggregation to augment the student pixels with richer contextual information and then aligns the augmented student pixels with the teacher pixels. Experiments on four challenging semantic segmentation datasets demonstrate that CIRKDV2 outperforms the state-of-the-art segmentation distillation methods. We hope our work can inspire future research to explore more advanced segmentation KD works towards global relational modeling and contextual feature distillation.
	
	\section*{Acknowledgment}
	This work was supported by the Youth Key Deployment Project of the Chinese Academy of Sciences under Grant No. GFQN-2026-34, the National Natural Science Foundation of China under Grant Nos. 62676389, 62406312, 62476264, 62606510, 62606511, and the Beijing Natural Science Foundation under Grant No. QG26011, the Postdoctoral Fellowship Program of CPSF under Grant Number GZC20260905.

	{\small
		\bibliographystyle{IEEEtran}
		\bibliography{IEEEbib}
	}

	
	\section{Biography Section}
	\vspace{-40pt}
	\begin{IEEEbiography}[{\includegraphics[width=1in,height=1.25in,clip,keepaspectratio]{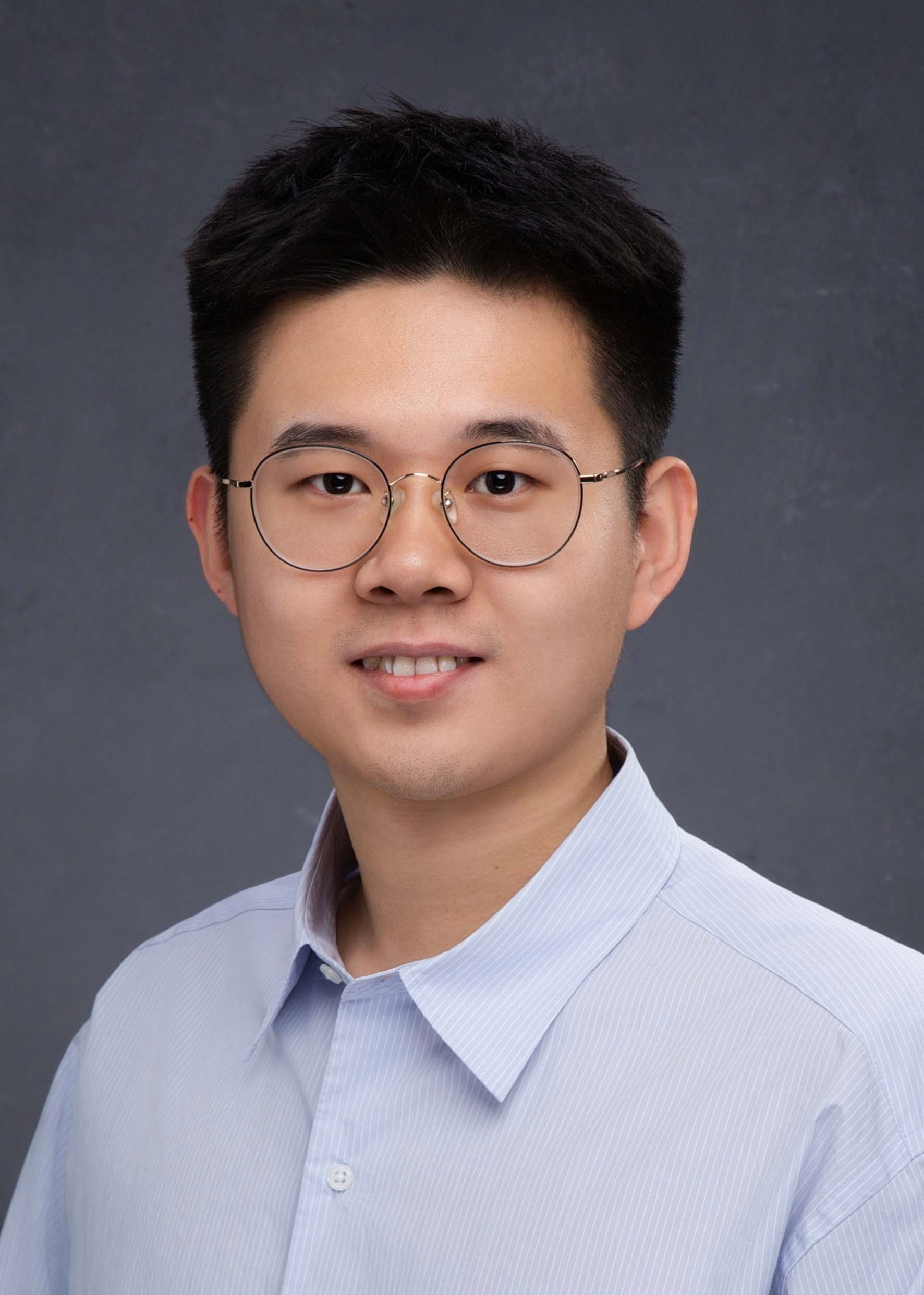}}]{Chuanguang Yang}
		received the B.E. degree from the Shandong Normal University, Jinan, China, in 2018, and the Ph.D. degree from the Institute of Computing Technology, Chinese Academy of Sciences, Beijing, China, in 2023. He is currently an associate professor at the Institute of Computing Technology, Chinese Academy of Sciences, Beijing, China. His research interests include knowledge distillation, model compression, and computer vision.
	\end{IEEEbiography}
	
	\vspace{-40pt}
	\begin{IEEEbiography}[{\includegraphics[width=1in,height=1.25in,clip,keepaspectratio]{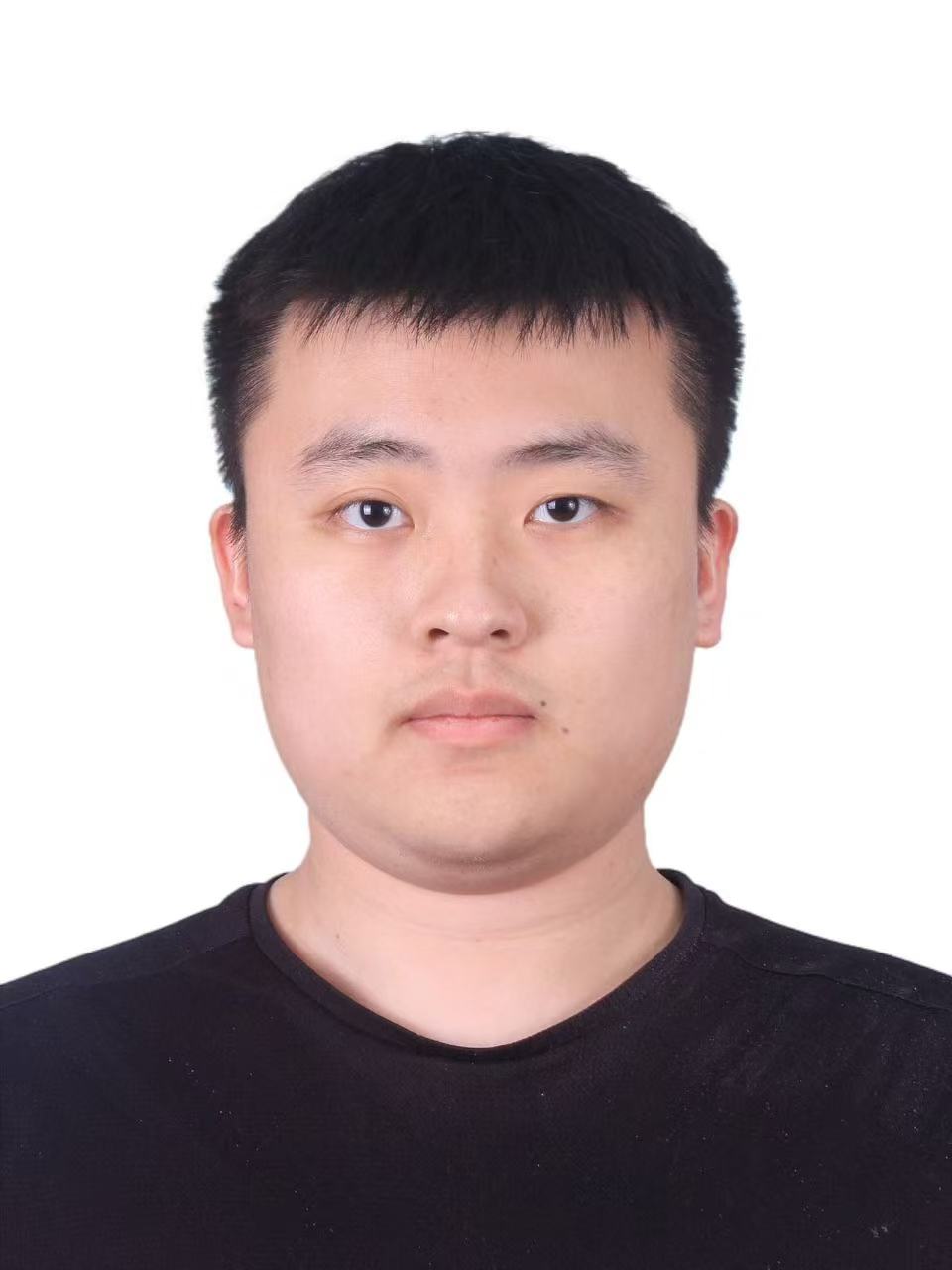}}]{Yu Wang}
		received the B.E. degree
		from Capital Normal University, Beijing, China,
		in 2022. He is currently pursuing a Ph.D. degree
		with the Institute of Computing Technology, Chinese Academy of Sciences, China. His research interests include knowledge distillation and visual generation.
	\end{IEEEbiography}

	\vspace{-40pt}
	\begin{IEEEbiography}[{\includegraphics[width=1in,height=1.25in,clip,keepaspectratio]{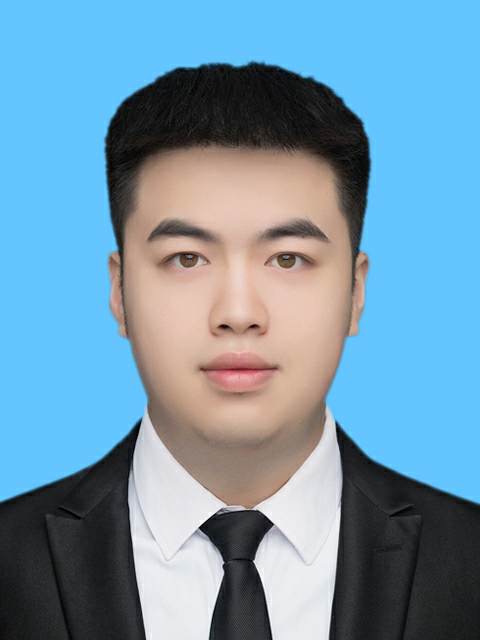}}]
		{Chengqing Yu} received the B.S. degree in Transport Equipment and Control Engineering from Central South University, Changsha, China, in 2019, the M.S. degree in Traffic and Transportation Engineering from the same university in 2022, and the Ph.D. degree from the Institute of Computing Technology, Chinese Academy of Sciences, Beijing, China, in 2026. He is currently an assistant professor with the Institute of Computing Technology, Chinese Academy of Sciences. His main research interests include deep learning and time series forecasting.
	\end{IEEEbiography}

	\vspace{-40pt}
	\begin{IEEEbiography}[{\includegraphics[width=1in,height=1.25in,clip,keepaspectratio]{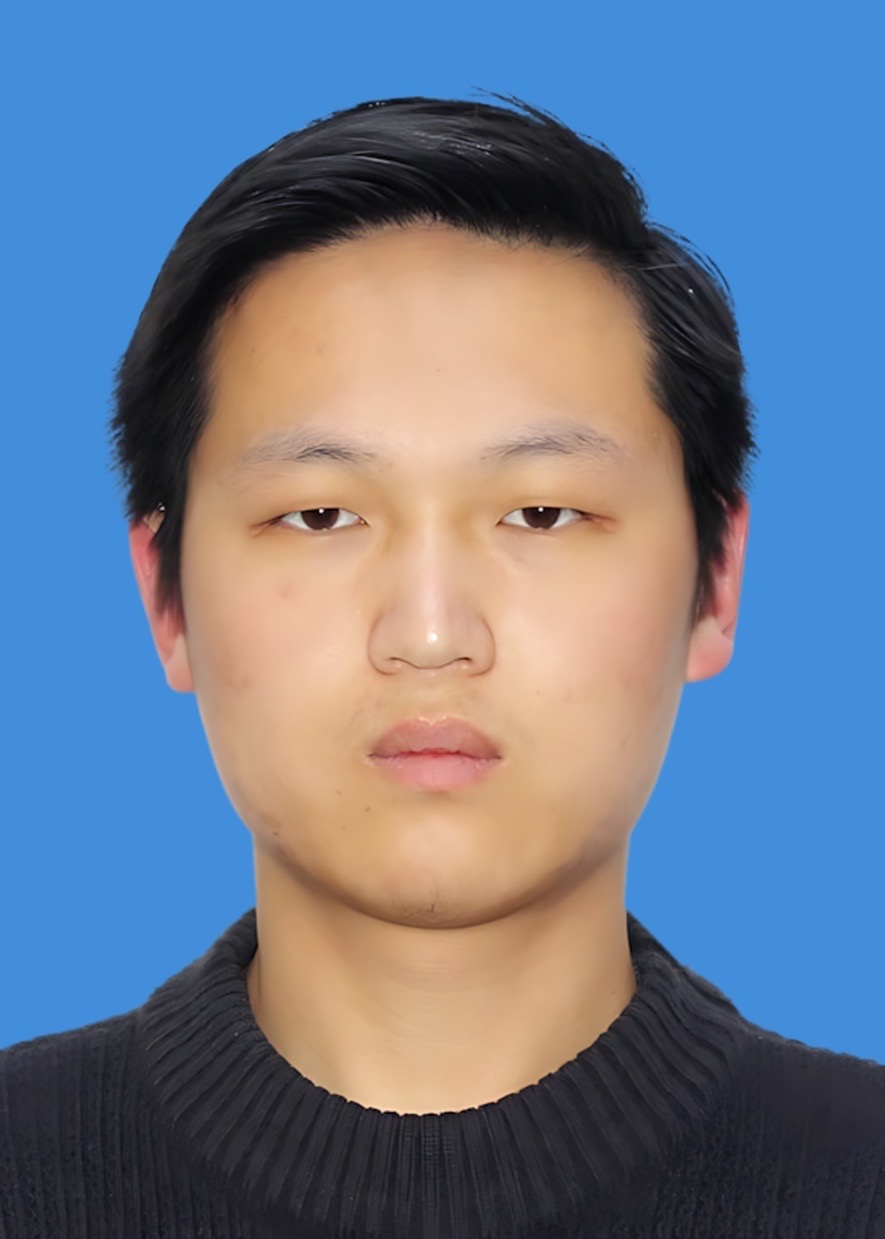}}]
		{Xinqiang Yu} received a B.E. degree from Liaoning Technical University, Liaoning, China, in 2020, and an M.E. degree from the Chinese Academy of Sciences, Beijing, China, in 2024. He is currently pursuing a Ph.D. degree with the Institute of Automation, Chinese Academy of Sciences, China. He has published papers on some prestigious refereed conferences such as CVPR, ICCV, AAAI and CoRL etc. His research interests include 3D computer vision, embodied AI and dexterous manipulation.
	\end{IEEEbiography}
	
	\vspace{-40pt}
	\begin{IEEEbiography}[{\includegraphics[width=1in,height=1.25in,clip,keepaspectratio]{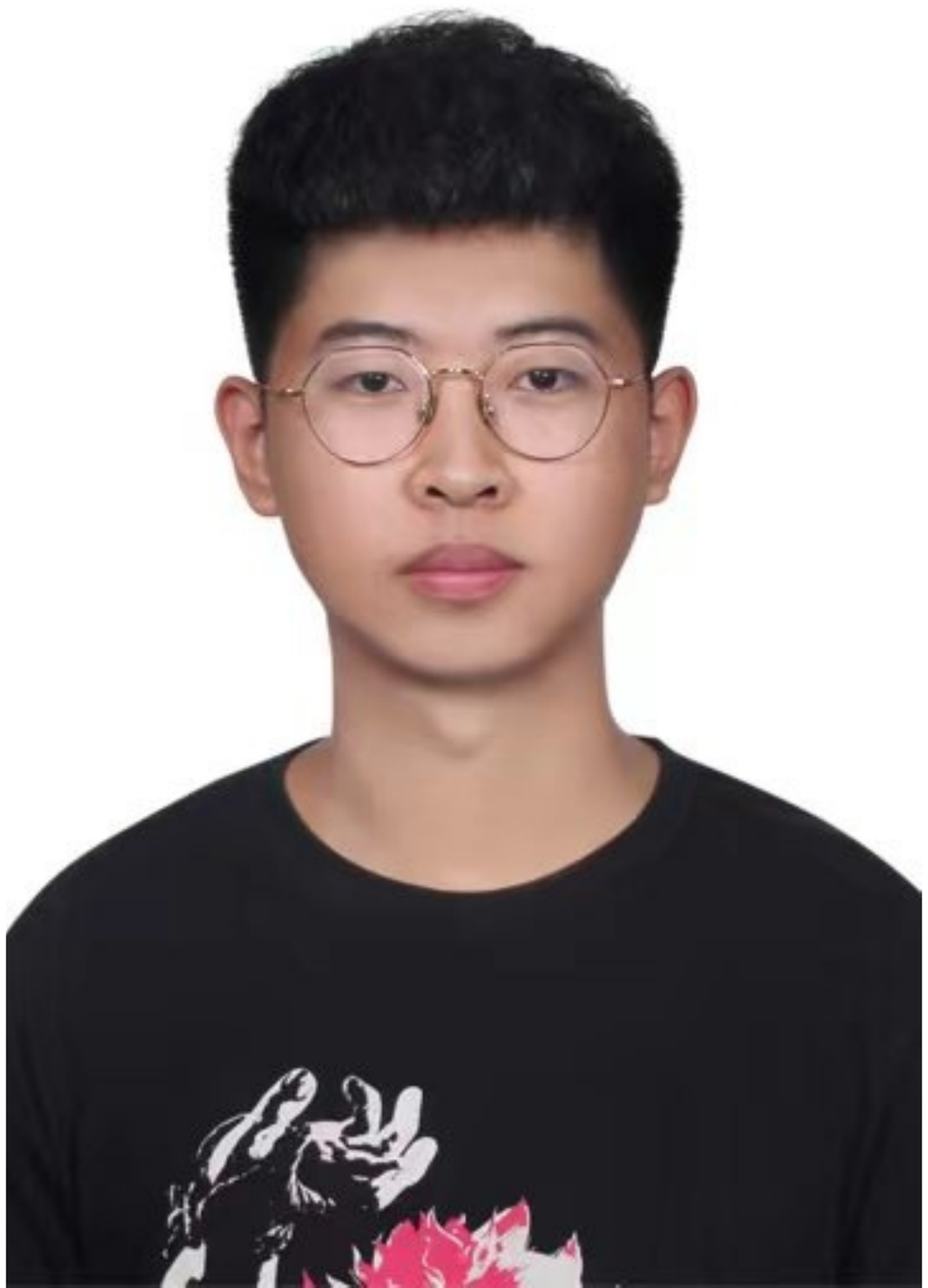}}]{Weilun Feng}
		received the B.E. degree from Beihang University, Beijing, China, in 2024. He is currently working toward the Master degree with the Institute of Computing Technology, Chinese Academy of Sciences, China. He has published papers on some prestigious refereed conferences such as ICML, CVPR, and AAAI etc. His research interests include model quantization, knowledge distillation, and visual generation.
	\end{IEEEbiography}
	
	\vspace{-30pt}
	
	\begin{IEEEbiography}[{\includegraphics[width=1in,height=1.25in,clip,keepaspectratio]{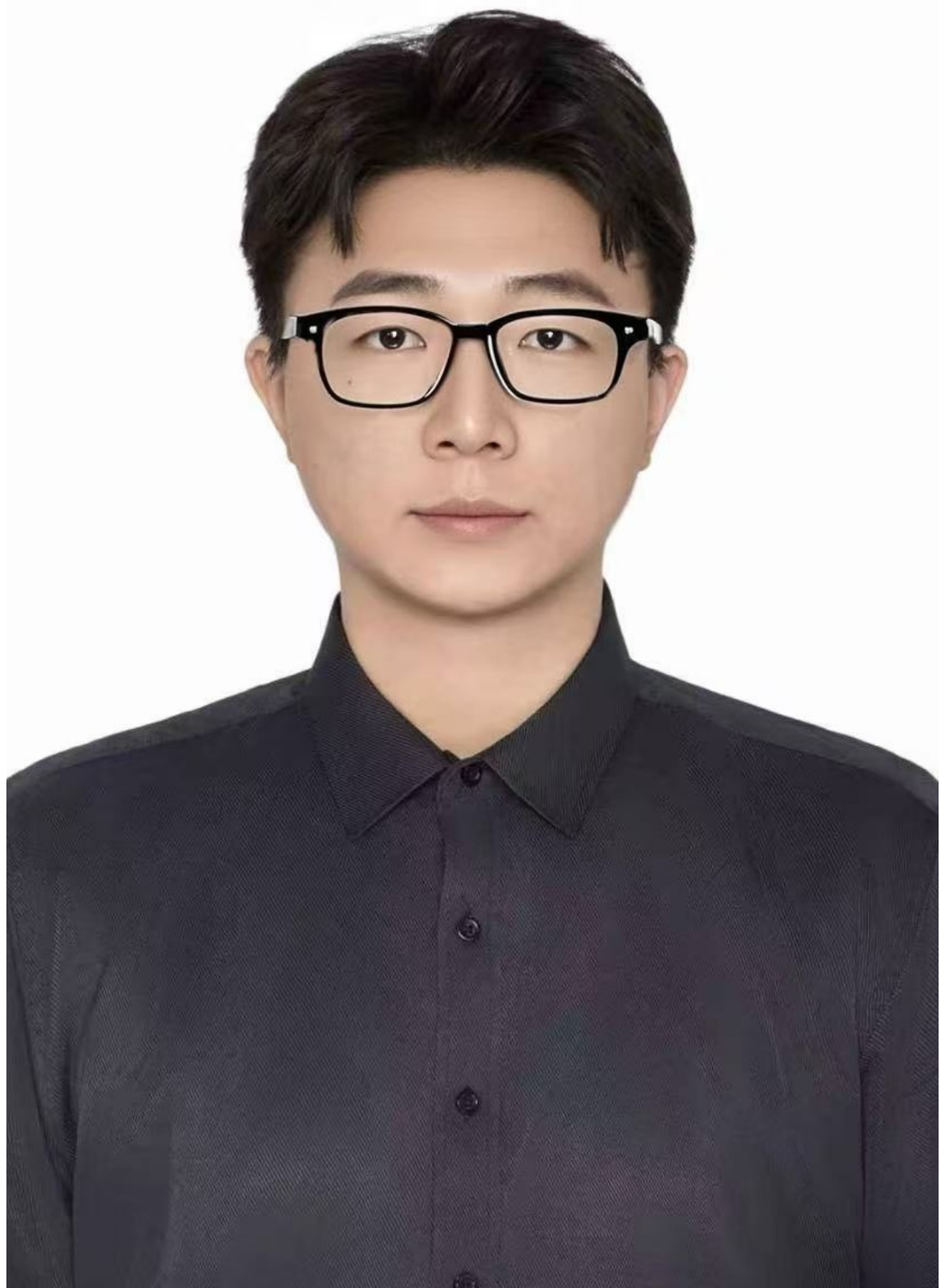}}]{Yuqi Li}
		is currently interning at the Institute of Computing Technology, Chinese Academy of Sciences. He graduated with a Bachelor's degree from Southwest University in 2023. His research interests lie in the field of computer vision and multimodal learning.
	\end{IEEEbiography}
	
	\vspace{-30pt}
	
	\begin{IEEEbiography}[{\includegraphics[width=1in,height=1.25in,clip,keepaspectratio]{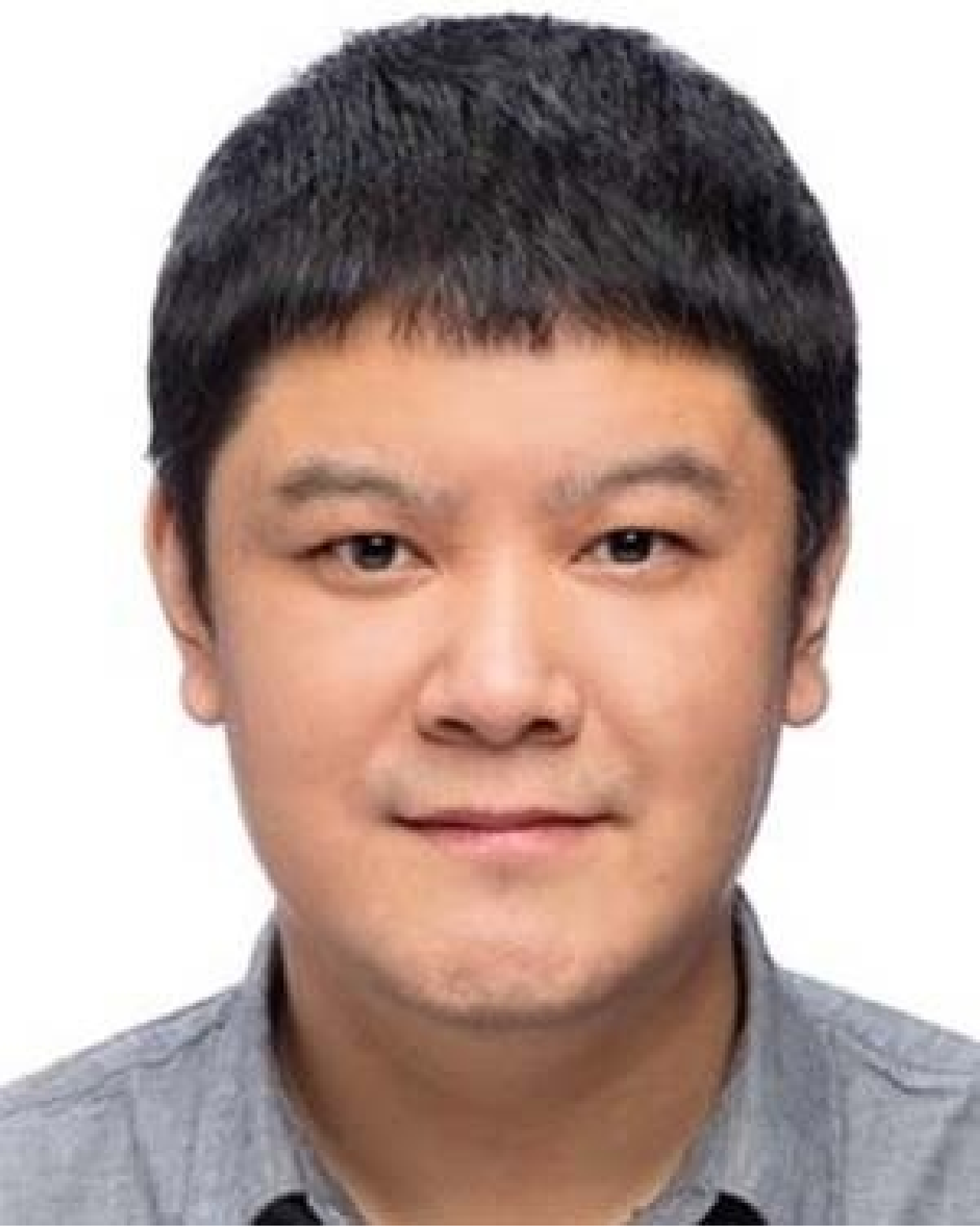}}]{Zhulin An}
		received B.Eng. and M.Eng. degrees in Computer Science from Hefei University of Technology, Hefei, China, in 2003 and 2006, respectively, and a Ph.D. degree in Computer Communication from the Chinese Academy of Sciences, Beijing, China, in 2010. He is currently an associate researcher at the Institute of Computing Technology, Chinese Academy of Sciences, Beijing, China. His research interests include the optimization of deep neural networks and lifelong learning.
	\end{IEEEbiography}

	\vspace{-30pt}
	\begin{IEEEbiography}[{\includegraphics[width=1in,height=1.25in,clip,keepaspectratio]{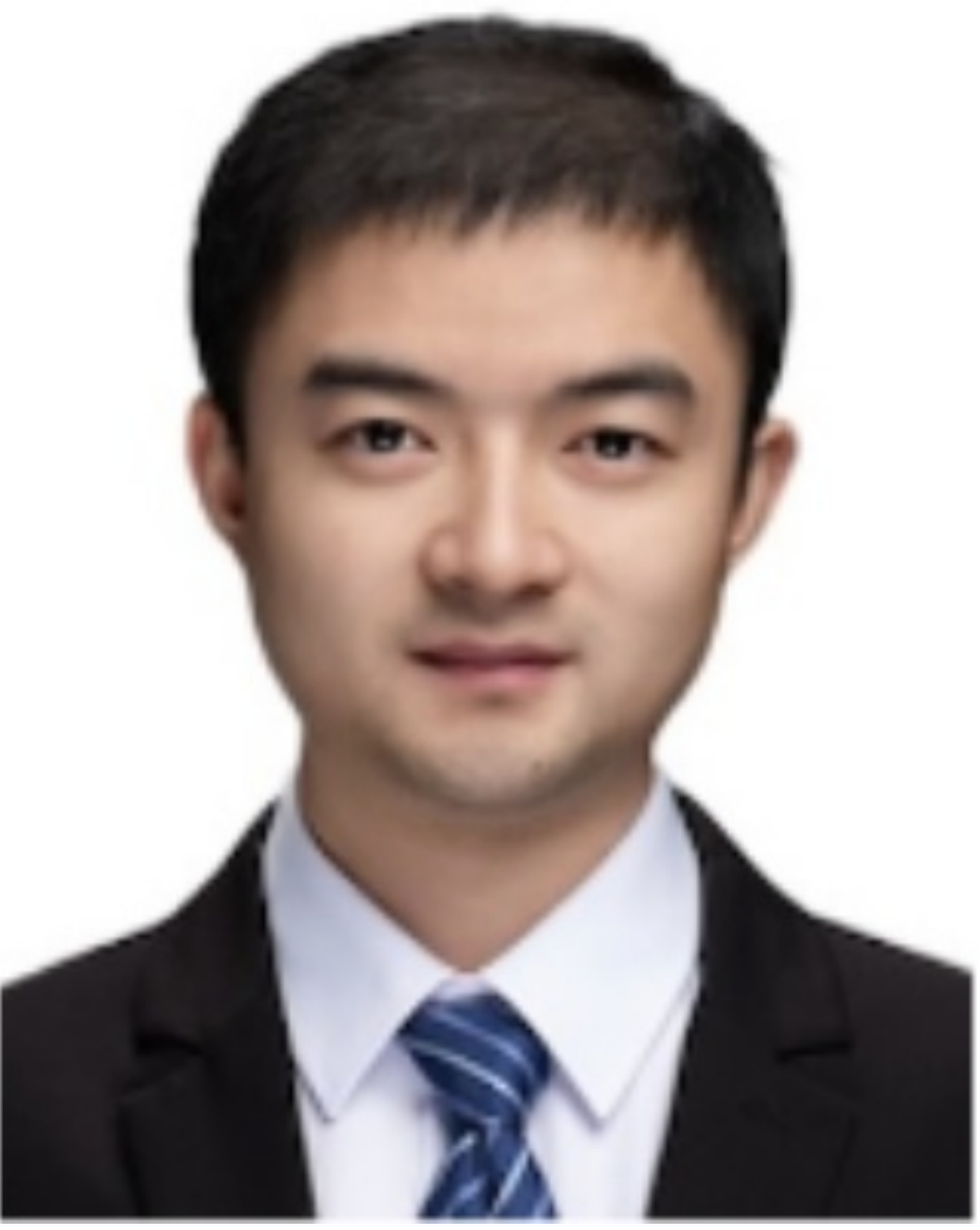}}]{Libo Huang}
		received a B.S. degree in Mathematics from Jiangxi Normal University, Jiangxi, China, in 2016, and a Ph.D. degree in Information Engineering from the Guangdong University of Technology, Guangdong, China, in 2021. He was a joint Ph.D. student at Brunel University, London, UK, from 2019 to 2020. He is currently an assistant researcher at the Institute of Computing Technology, Chinese Academy of Sciences, Beijing, China. His research interests include machine learning theories and deep learning applications.
	\end{IEEEbiography}
	
	\vspace{-30pt}
	\begin{IEEEbiography}[{\includegraphics[width=1in,height=1.25in,clip,keepaspectratio]{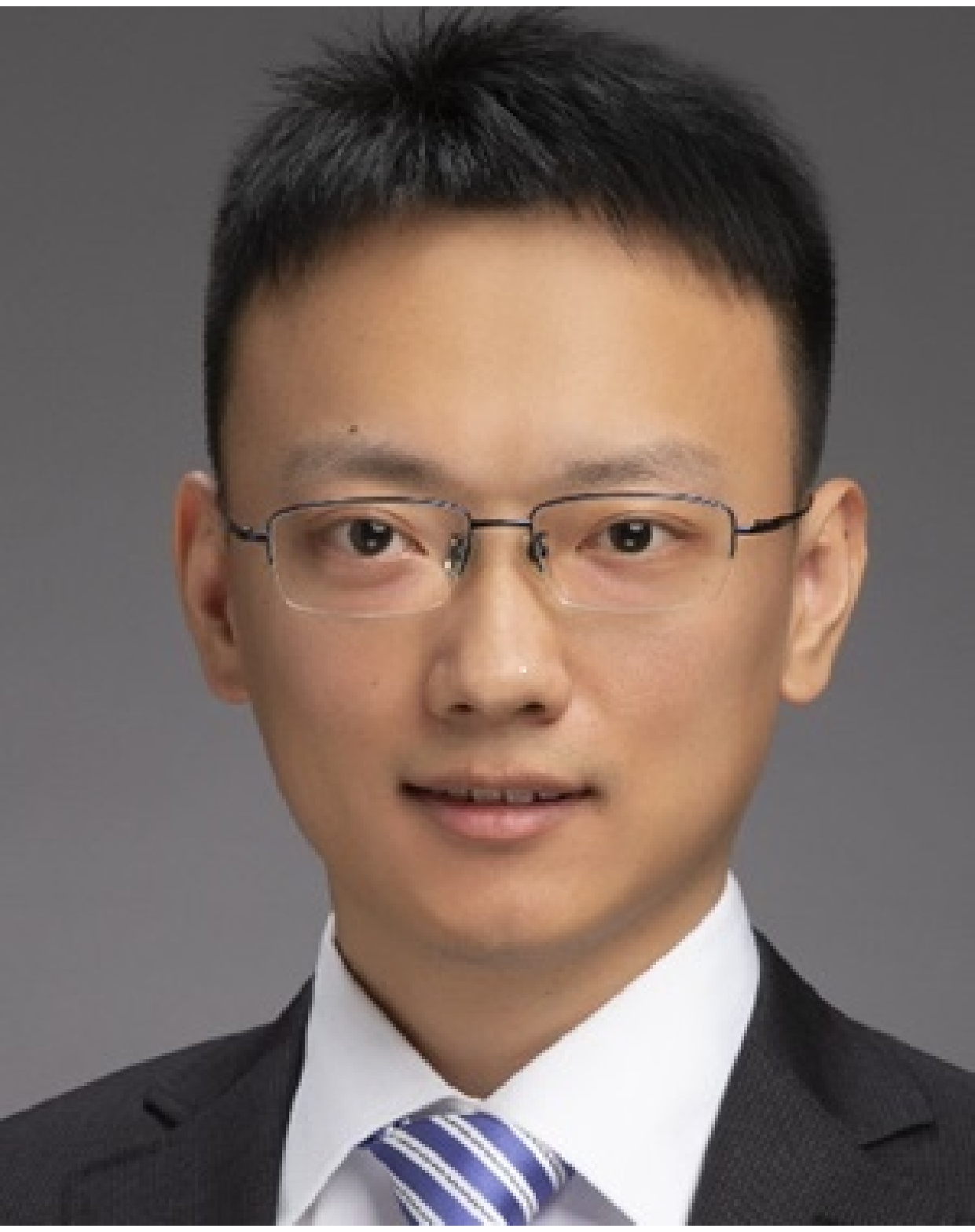}}]{Boyu Diao}
		is an Associate Professor at the Institute of Computing Technology, Chinese Academy of Sciences; He received his Ph.D. in Computer Architecture from ICT CAS in 2018 and his B.Eng. in Computer Science from Beijing Institute of Technology in 2012. He leads a group towards Efficient Machine Learning for Edge Intelligence Systems with main research interests in algorithm and hardware co-design for efficient deep learning, deep learning compilers, and on-device learning systems.
	\end{IEEEbiography}
	
	\vspace{-30pt}
	
	\begin{IEEEbiography}[{\includegraphics[width=1in,height=1.25in,clip,keepaspectratio]{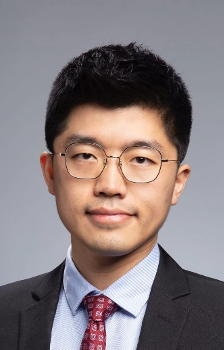}}]
		{Fei Wang} received the B.S. degree in computer science from the Beijing Institute of Technology, Beijing, China, in 2011. He received the Ph.D. degree in computer architecture from Institute of Computing Technology, Chinese Academy of Sciences in 2017. From 2017 to 2020, he was a research assistant with the Institute of Computing Technology, Chinese Academy of Sciences. Since 2020, he has been working as associate professor in Institute of Computing Technology, Chinese Academy of Sciences. His main research interests include spatiotemporal data mining, Information fusion, and graph neural networks.
	\end{IEEEbiography}
	
	\vspace{-40pt}
	\begin{IEEEbiography}[{\includegraphics[width=1in,height=1.25in,clip,keepaspectratio]{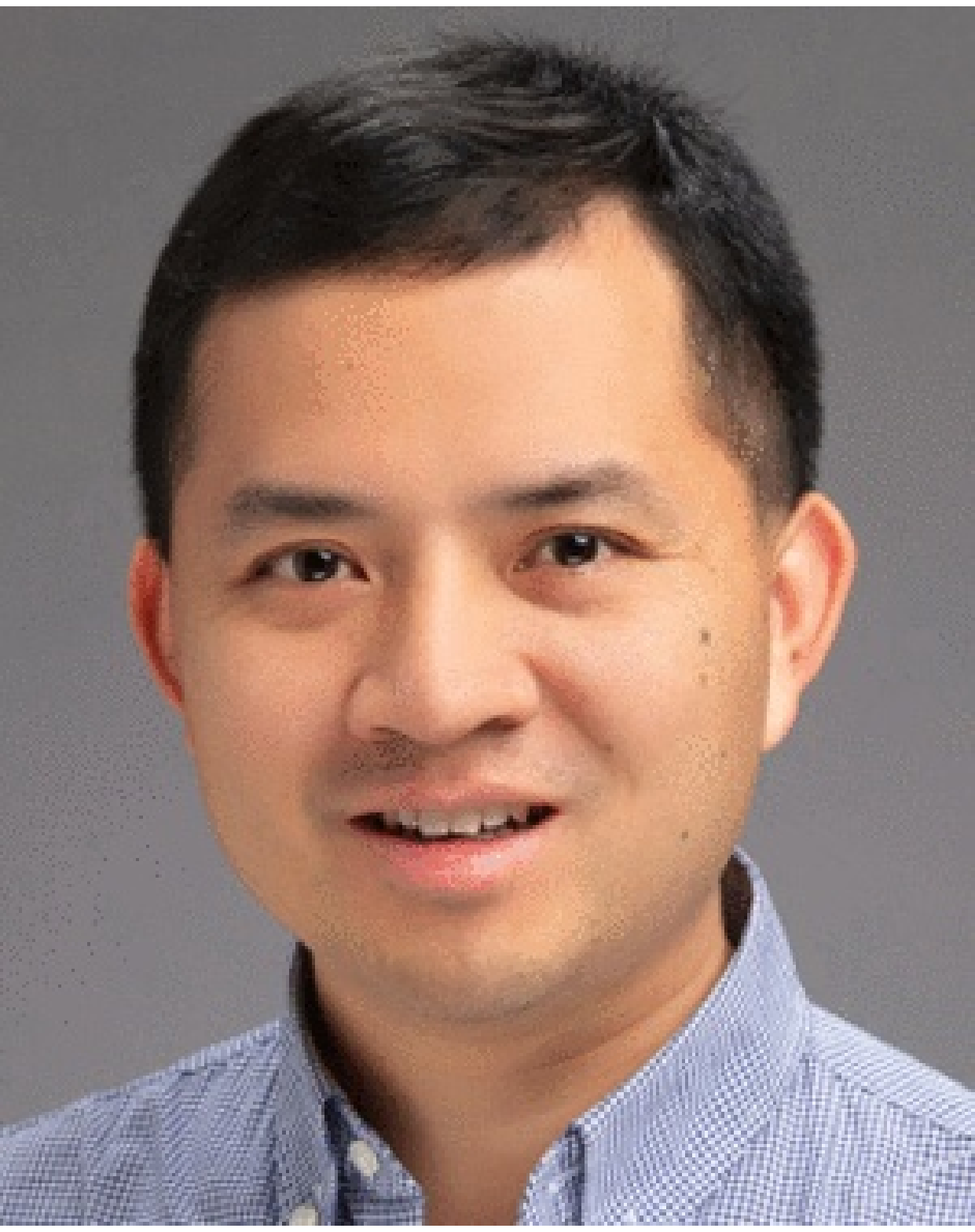}}]{Fuzhen Zhuang}
		(Member, IEEE) received the B.E. degree from the College of Computer Science, Chongqing University, Chongqing, China, in 2006, and the Ph.D. degree in computer science from the Institute of Computing Technology, Chinese Academy of Sciences, Beijing, China, in 2011. He is currently a full professor with the Institute of Artificial Intelligence, Beihang University. He has published more than 200 papers in many prestigious refereed journals and conference proceedings. His research interests include transfer learning, machine learning, data mining, multitask learning, knowledge graph, and recommendation systems. 
	\end{IEEEbiography}

	\vspace{-30pt}
	\begin{IEEEbiography}[{\includegraphics[width=1in,height=1.25in,clip,keepaspectratio]{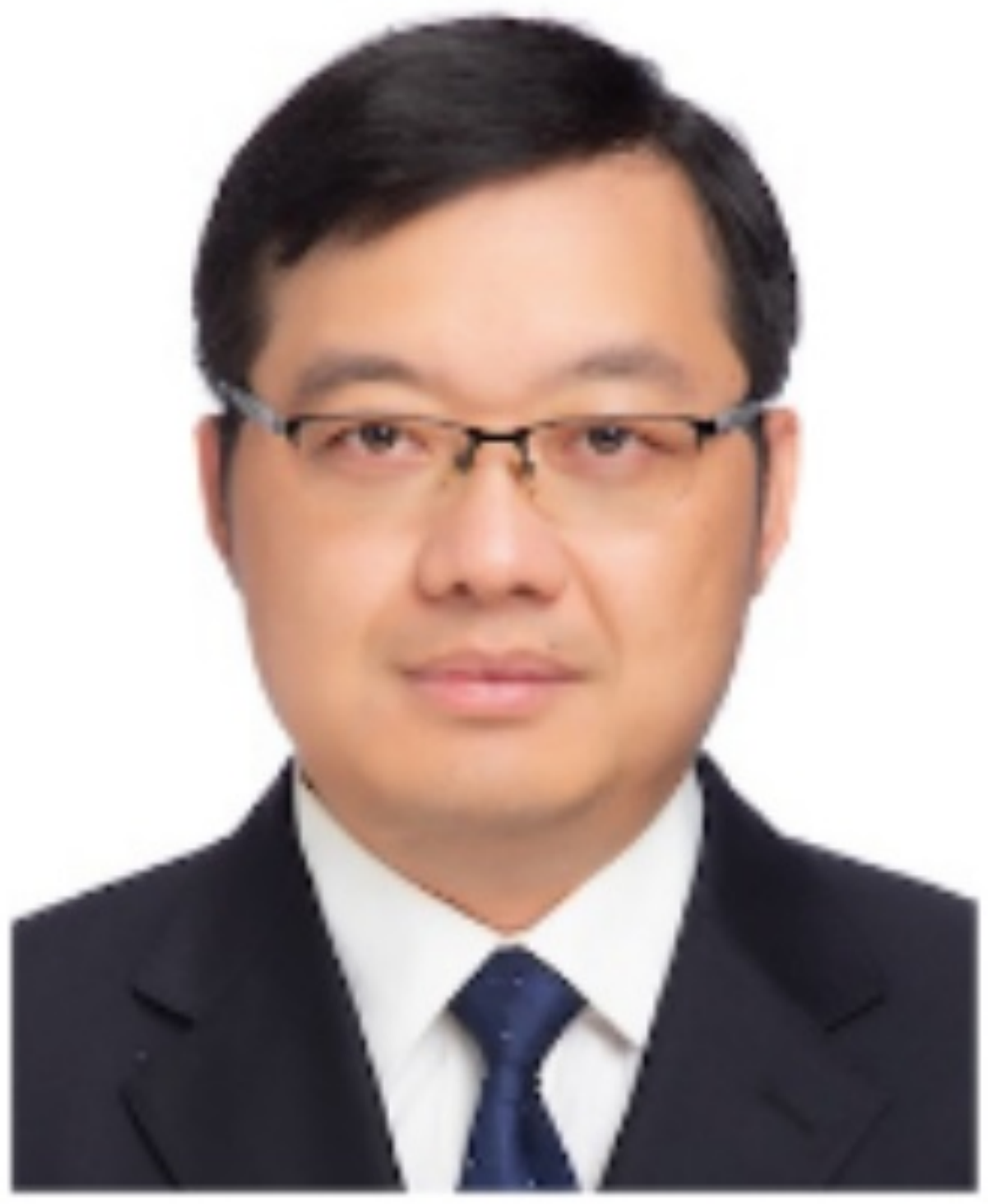}}]{Yongjun Xu}
		received a B.Eng. degree in Computer Communication from the Xi’an Institute of Posts and Telecoms, Xi'an, China, in 2001, and a Ph.D. degree in Computer Communication from the Institute of Computing Technology, Chinese Academy of Sciences, Beijing, China, in 2006. He is currently a researcher at the Institute of Computing Technology, Chinese Academy of Sciences, Beijing, China. His research interests include artificial intelligence systems and big data processing.
	\end{IEEEbiography}
	
	\vspace{-40pt}
	\begin{IEEEbiography}[{\includegraphics[width=1in,height=1.25in,clip,keepaspectratio]{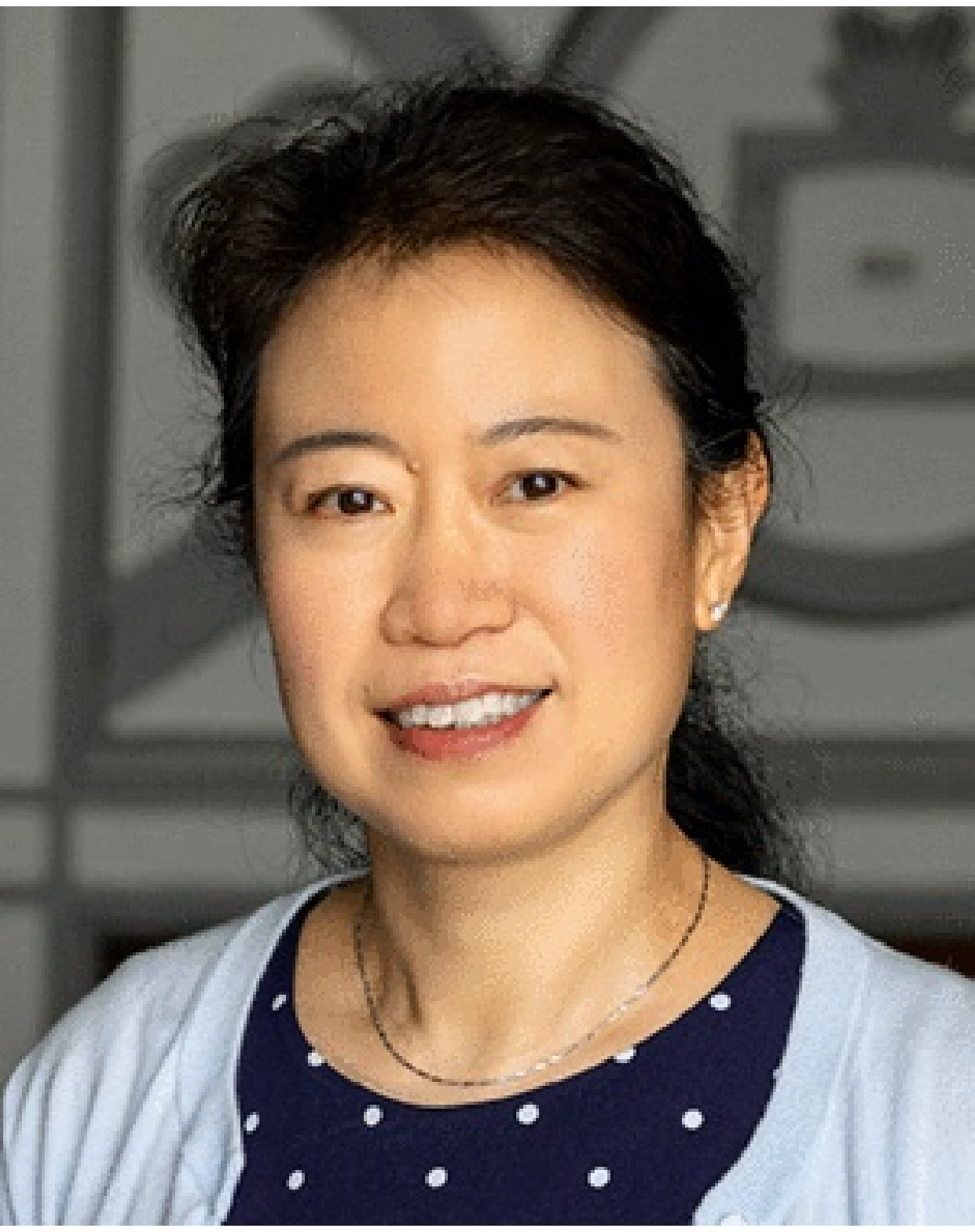}}]{Yingli Tian}
		(M'99--SM'01--F'18) received her B.S. and M.S. degrees from Tianjin University, China, in 1987 and 1990, respectively, and the Ph.D. degree from the Chinese University of Hong Kong in 1996. After holding a faculty position at the National Laboratory of Pattern Recognition, Chinese Academy of Sciences, Beijing, she joined Carnegie Mellon University in 1998 as a postdoctoral fellow at the Robotics Institute. From 2001 to 2008, she worked as a research staff member at the IBM T. J. Watson Research Center. She is one of the inventors of the IBM Smart Surveillance Solutions. Currently, she is a CUNY Distinguished Professor in the Department of Electrical Engineering at the City College and the Department of Computer Science at the Graduate Center, the City University of New York. Her research focuses on a wide range of computer vision problems, including scene understanding, medical imaging analysis, human behavior analysis, facial expression recognition, and assistive technology. She is a Fellow of IEEE, IAPR, AAAS, and AAIA.
	\end{IEEEbiography}
	
	\vspace{-30pt}
	\begin{IEEEbiography}[{\includegraphics[width=1in,height=1.25in,clip,keepaspectratio]{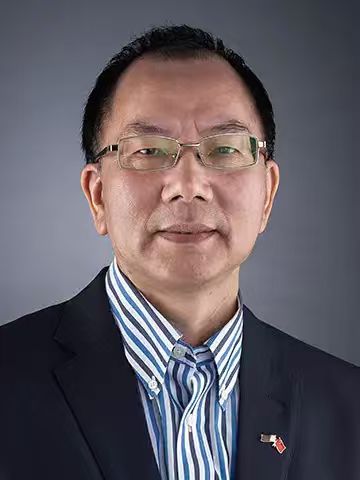}}]{Tingwen Huang}
		(Fellow, IEEE) received the B.S.
		degree in mathematics from Southwest University, Chongqing, China, in 1990, the M.S. degree in applied mathematics from Sichuan University, Chengdu, China, in 1993, and the Ph.D. degree in applied mathematics from Texas A\&M University at College Station, College Station, TX, USA, in
		2002.
		
		Following his graduation, he worked as a Visiting Assistant Professor with Texas A\&M University at College Station. Then, he joined Texas A\&M
		University at Qatar, Doha, Qatar, as an Assistant Professor in August 2003, then he was promoted to Professor in 2013. His focus areas for research interests include computational intelligence, smart grid, dynamical systems, optimization and control, and cybersecurity.
	\end{IEEEbiography}
	
	\vspace{-30pt}
	\begin{IEEEbiography}[{\includegraphics[width=1in,height=1.25in,clip,keepaspectratio]{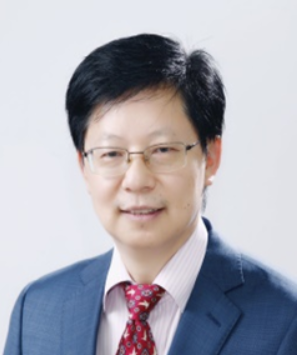}}]{Yongduan Song}
		  (Fellow, IEEE) received the Ph.D. degree in electrical and computer engineering from Tennessee Technological University, Cookeville, TN, USA, in 1992. He was a tenured Full Professor with the North Carolina A \& T State University, Greensboro, NC, USA, from 1993 to 2008. He was one of the six Langley Distinguished Professors of National Institute of Aerospace (NIA), Hampton, VA, USA and the Founding Director of Cooperative Systems at NIA. He is currently the Chair Professor with the School of Automation, Chongqing University, Chongqing, China. His current research interests include intelligent systems, guidance navigation and control, and bio-inspire adaptive and cooperative systems. Dr. Song has served/been serving as an Associate Editor for several prestigious international journals, including the IEEE Transactions on Automatic Control, IEEE Transactions on Neural Networks and Learning Systems, IEEE Transactions on Intelligent Transportation Systems, and IEEE Transactions on Systems, Man, and Cybernetics. He is currently the Editor-in-Chief of the IEEE Transactions on Neural Networks and Learning Systems.
	\end{IEEEbiography}
	\vfill
	
\end{document}